\documentclass{article}

\usepackage{microtype}
\usepackage{graphicx}
\usepackage{booktabs} 

\usepackage{hyperref}

\usepackage[accepted]{icml2018}

\usepackage{subcaption}		

\renewcommand{\algorithmicrequire}{\textbf{Input:}}
\renewcommand{\algorithmicensure}{\textbf{Output:}}

\usepackage{amsmath}
\usepackage{amsfonts}								
\usepackage{dsfont}            			
\newcommand{\argmin}{\operatorname*{argmin}}	
\newcommand{\Lap}{\operatorname{Lap}}		
\newcommand{\supp}{\operatorname{supp}}	
\newcommand{\calA}{\mathcal{A}}			
\newcommand{\calH}{\mathcal{H}}			
\newcommand{\calN}{\mathcal{N}}			
\newcommand{\calO}{\mathcal{O}}			
\newcommand{\calX}{\mathcal{X}}			
\newcommand{\IE}{\mathbb{E}}  			
\newcommand{\IN}{\mathbb{N}}  			
\newcommand{\IP}{\mathbb{P}}  			
\newcommand{\IR}{\mathbb{R}}  			

\usepackage{amsthm}
\theoremstyle{plain}
\newtheorem{theorem}{Theorem}
\newtheorem{corollary}[theorem]{Corollary}
\newtheorem{lemma}[theorem]{Lemma}
\newtheorem{proposition}[theorem]{Proposition}
\theoremstyle{remark}
\newtheorem{remark}[theorem]{Remark}
\theoremstyle{definition}
\newtheorem{definition}[theorem]{Definition}

\usepackage{thmtools}
\declaretheoremstyle[notefont=\bfseries,notebraces={}{},headpunct={},postheadspace=1em]{mystyle}
\declaretheorem[style=mystyle,numbered=no,name=Theorem]{hthm} 		
\declaretheorem[style=mystyle,numbered=no,name=Proposition]{hprop} 	

\icmltitlerunning{Differentially Private Database Release via Kernel Mean Embeddings}

\begin{document}

\twocolumn[
\icmltitle{Differentially Private Database Release via Kernel Mean Embeddings}



\icmlsetsymbol{equal}{*}

\begin{icmlauthorlist}
\icmlauthor{Matej Balog}{ei,cam}
\icmlauthor{Ilya Tolstikhin}{ei}
\icmlauthor{Bernhard Sch\"olkopf}{ei}
\end{icmlauthorlist}

\icmlaffiliation{ei}{MPI-IS, T{\"u}bingen, Germany}
\icmlaffiliation{cam}{University of Cambridge, UK}

\icmlcorrespondingauthor{Matej Balog}{first.surname@gmail.com}

\icmlkeywords{Machine Learning, ICML, kernels, RKHS, differential privacy, synthetic database}

\vskip 0.3in
]



\printAffiliationsAndNotice{}  

\begin{abstract}
We lay theoretical foundations for new database release mechanisms that allow third-parties to construct consistent estimators of population statistics, while ensuring that the privacy of each individual contributing to the database is protected. The proposed framework rests on two main ideas. First, releasing (an estimate of) the kernel mean embedding of the data generating random variable instead of the database itself still allows third-parties to construct consistent estimators of a wide class of population statistics. Second, the algorithm can satisfy the definition of differential privacy by basing the released kernel mean embedding on entirely synthetic data points, while controlling accuracy through the metric available in a Reproducing Kernel Hilbert Space. We describe two instantiations of the proposed framework, suitable under different scenarios, and prove theoretical results guaranteeing differential privacy of the resulting algorithms and the consistency of estimators constructed from their outputs.
\end{abstract}

\section{Introduction}
\label{sec:introduction}

We aim to contribute to the body of research on the trade-off between releasing datasets from which publicly beneficial statistical inferences can be drawn, and between protecting the privacy of individuals who contribute to such datasets. Currently the most successful formalisation of protecting user privacy is provided by \emph{differential privacy}~\cite{dwork_algorithmic_2014}, which is a \emph{definition} that any algorithm operating on a database may or may not satisfy. An algorithm that does satisfy the definition ensures that a particular individual does not lose too much privacy by deciding to contribute to the database on which the algorithm operates.

While differentially private algorithms for releasing entire databases have been studied previously~\cite{blum_learning_2008,wasserman_statistical_2008,zhou_differential_2009}, most algorithms focus on releasing a privacy-protected version of a particular summary statistic, or of a statistical model trained on the private dataset. In this work we revisit the more difficult \emph{non-interactive}, or \emph{offline} setting, where the database owner aims to release a privacy-protected version of the entire database without knowing what statistics third-parties may wish to compute in the future.

In our new framework we propose to use the kernel mean embedding~\cite{smola_hilbert_2007} as an intermediate representation of a database.
It is (1) sufficiently rich in the sense that it captures a wide class of statistical properties of the data, while at the same time (2) it lives in a Reproducing Kernel Hilbert Space (RKHS), where it can be handled mathematically in a principled way and privacy-protected in a unified manner, independently of the type of data appearing in the database.
Although kernel mean embeddings are functions in an abstract Hilbert space, in practice they can be (at least approximately) represented using a possibly weighted set of data points in input space (i.e.~a set of database rows).
The privacy-protected kernel mean embedding is released to the public in this representation, however, using synthetic datapoints instead of the private ones. As a result, our framework can be seen as leading to \emph{synthetic database} algorithms.

We validate our approach by instantiating two concrete algorithms and proving that they output consistent estimators of the true kernel mean embedding of the data generating process, while satisfying the definition of differential privacy. The consistency results ensure that third-parties can carry out a wide variety of statistically founded computation on the released data, such as constructing consistent estimators of population statistics, estimating the Maximum Mean Discrepancy (MMD) between distributions, and two-sample testing~\citep{gretton_kernel_2012}, or using the data in the kernel probabilistic programming framework for random variable arithmetics~\citep[Section 3]{scholkopf_computing_2015, scibior_consistent_2016}, repeatedly and unlimitedly without being able to, or having to worry about, violating user privacy.

One of our algorithms is especially suited to the interesting scenario where a (small) subset of a database has already been published. This situation can arise in a wide variety of settings, for example, due to weaker privacy protections in the past, due to a leak, or due to the presence of an incentive, financial or otherwise, for users to publish their data. In such a situation our algorithm provides a principled approach for reweighting the public data in such a way that the accuracy of statistical inferences on this dataset benefits from the larger sample size (including the private data), while maintaining differential \mbox{privacy for the undisclosed data}.

In summary, the contributions of this paper are:
\begin{enumerate}
	\item A new framework for designing database release algorithms with the guarantee of differential privacy. The framework uses kernel mean embeddings as intermediate database representations, so that the RKHS metric can be used to control accuracy of the released synthetic database in a principled manner (Section~\ref{sec:framework}).
	\item Two instantiations of our framework in the form of two synthetic database algorithms, with proofs of their consistency, convergence rates and differential privacy, as well as basic empirical illustrations of their performance on synthetic datasets (Sections~\ref{sec:perturb_in_synthetic_subspace} and~\ref{sec:perturb_in_random_features_RKHS}).
\end{enumerate}

\section{Background}
\label{sec:background}

\subsection{Differential Privacy}
\label{sec:background:differential_privacy}

\begin{definition}[\citealp{dwork_differential_2006}]
	For $\varepsilon > 0$, $\delta \geq 0$, algorithm $\calA$ is said to be $(\varepsilon, \delta)$-differentially private if for all neighbouring databases $D \sim D'$ (differing in at most one element) and all measurable subsets $S$ of the co-domain of $\calA$,
	\begin{equation}
	\IP\left( \calA(D) \in S \right)
	\leq e^{\varepsilon} \IP\left( \calA(D') \in S \right) + \delta
	.
	\end{equation}
\end{definition}

The parameter $\varepsilon$ controls the amount of information the algorithm can leak about an individual, while a positive $\delta$ allows the algorithm to produce an unlikely output that leaks more information, but only with probability up to $\delta$. This notion is sometimes called \emph{approximate} differential privacy; an algorithm that is $(\varepsilon, 0)$-differentially private is simply said to be $\varepsilon$-differentially private. Note that any non-trivial differentially private algorithm must be randomised; the definition asserts that the distribution of algorithm outputs is not too sensitive to changing one row in the database.

When the algorithm's output is a finite vector $\calA(D) \in \IR^J$, two standard random perturbation mechanisms for making this output differentially private are the \emph{Laplace} and \emph{Gaussian} mechanisms. As the perturbation needs to mask the contribution of each individual entry of the database $D$, the scale of the added noise is closely linked to the notion of \emph{sensitivity}, measuring how much the algorithm's output can change due to changing a single data point:
\begin{align}
\Delta_1
&:=
\sup_{D \sim D'} \left\| \calA(D) - \calA(D') \right\|_1
,
\\
\Delta_2
&:=
\sup_{D \sim D'} \left\| \calA(D) - \calA(D') \right\|_2
.
\end{align}
The Laplace mechanism adds i.i.d.~$\Lap(\Delta_1 / \varepsilon)$ noise to each of the $J$ coordinates of the output vector and ensures pure $\varepsilon$-differential privacy, while the Gaussian mechanism adds i.i.d.~$\calN(0, \sigma^2)$ noise to each coordinate, where $\sigma^2 > 2 \Delta_2^2 \ln(1.25 / \delta) / \varepsilon^2$, and ensures $(\varepsilon, \delta)$-differential privacy. Applying these mechanisms thus requires computing (an upper bound on) the relevant sensitivity.

Differential privacy is preserved under post-processing: if an algorithm $\calA$ is $(\varepsilon, \delta)$-differentially private, then so is its sequential composition $\mathcal{B}(\calA(\cdot))$ with any other algorithm $\mathcal{B}$ that does not have direct or indirect access to the private database $D$~\cite{dwork_algorithmic_2014}.

\subsection{Kernels, RKHS, and Kernel Mean Embeddings}
\label{sec:background:kernels}

A kernel on a non-empty set (data type) $\calX$ is a binary positive-definite function $k(\cdot, \cdot) : \calX \times \calX \to \IR$.  Intuitively it can be thought of as expressing the similarity between any two elements in $\calX$. The literature on kernels is vast and their properties are well studied~\cite{scholkopf_learning_2001}; many kernels are known for a large variety of data types such as vectors, strings, time series, graphs, etc, and kernels can be composed to yield valid kernels for composite data types (e.g.~the type of a database row containing both numerical and string data).

The \emph{kernel mean embedding} (KME) of an $\calX$-valued random variable $X$ in the RKHS is the function $\mu_X^k : \calX \to \IR$, $y \mapsto \mathbb{E}_X[k(X, y)]$, defined whenever $E_X[\sqrt{k(X, X)}] < \infty$~\cite{smola_hilbert_2007}. Several popular kernels have been proved to be \emph{characteristic}~\cite{FukGreSunSch08}, in which case the map $p_X \mapsto \mu_X^k$, where $p_X$ is the distribution of $X$, is injective. This means that no information about the distribution of $X$ is lost when passing to its KME $\mu_X^k$.

In practice, the KME of a random variable $X$ is approximated using a sample $x_1, \ldots, x_N$ drawn from $X$, which can be used to construct an \emph{empirical KME} $\hat{\mu}_X^k$ of $X$ in the RKHS: a function given by $y \mapsto \frac{1}{N} \sum_{n = 1}^N k(x_n, y)$.
When the $x_n$'s are i.i.d., under a boundedness condition $\hat{\mu}_X^k$ converges to the true KME $\mu^k_X$ at rate $\calO_p(N^{-1/2})$, independently of the dimension of $\calX$ \citep{LopMuaSchTol15}%
\footnote{The KME can be viewed as a smoothed version of the density, which is easier to estimate than the density itself; rates of nonparametric density estimation or statistical powers of two-sample or independence tests involving $p_X$ are known to necessarily degrade with growing dimension~\citep[Section 4.3]{tolstikhin2017minimax}.}%
.
Our approach relies on the metric of the RKHS in which these KMEs live. The RKHS $\calH_k$ is a space of functions, endowed with an inner product $\langle \cdot, \cdot \rangle_{\calH_k}$ that satisfies the \emph{reproducing} property $\langle k(x, \cdot), h \rangle = h(x)$ for all $x \in \calX$ and $h \in \calH_k$. The inner product induces a norm $\| \cdot \|_{\calH_k}$, which can be used to measure distances $\| \mu^k_X - \mu^k_Y \|_{\calH_k}$ between distributions of $X$ and $Y$. This can be exploited for various purposes such as two-sample tests~\citep{gretton_kernel_2012}, independence testing~\citep{gretton_measuring_2005}, or one can attempt to minimise this distance in order to match one distribution to another.

An example of such minimisation are \emph{reduced set methods}~\citep[Chap.~18]{burges_simplified_1996, scholkopf_learning_2001}, which replace a set of points $S = \{x_1, \ldots, x_N \} \subseteq \calX$ with a weighted set $R = \{ (z_1, w_1), \ldots, (z_M, w_M) \} \subseteq \calX \times \IR$ (of potentially smaller size), where the new points $z_m$ can, but need not equal any of the $x_n$s, such that the KME computed using the reduced set $R$ is close to the KME computed using the original set $S$, as measured by the RKHS norm:
\begin{align*}
&\left\| \mu^k_S - \mu^k_R \right\|_{\calH_k}
\nonumber
=
\left\|
\frac{1}{N} \sum_{n = 1}^N k(x_n, \cdot)
- \sum_{m = 1}^M w_m k(z_m, \cdot)
\right\|_{\mathcal{H}_k}
.
\end{align*}
Reduced set methods are usually motivated by the computational savings arising when $|R| < |S|$; we will invoke them mainly to replace a collection $S$ of private data points with a (possibly weighted) set $R$ of synthetic data points.

\section{Framework}
\label{sec:framework}

\subsection{Problem Formulation}
\label{sec:framework:problem_formulation}

Throughout this work, we assume the following setup. A database curator wishes to publicly release a database $D = \{ x_1, \ldots x_N \} \in \calX^N$ containing private data about $N$ individuals, with each data point (database row) $x_n$ taking values in a non-empty set $\calX$. The set $\calX$ can be arbitrarily rich, for example, it could be a product of Euclidean spaces, integer spaces, sets of strings, etc.; we only require availability of a kernel function $k : \calX \times \calX \to \IR$ on $\calX$. We assume that the $N$ rows $x_1, \ldots, x_N$ in the database $D$ can be thought of as i.i.d.~observations from some $\calX$-valued data-generating random variable $X$ (but see Section~\ref{sec:discussion} for a discussion about relaxing this assumption). The database curator, wishing to protect the privacy of individuals in the database, seeks a database release mechanism that satisfies the definition of $(\varepsilon, \delta)$-differential privacy, with $\varepsilon > 0$ and $\delta \geq 0$ given. The main purpose of releasing the database is to allow third parties to construct estimators of population statistics (i.e.~properties of the distribution of $X$), but it is not known at the time of release what statistics the third-parties will be interested in.

To lighten notation, henceforth we drop the superscript $k$ from KMEs (such as $\mu_X^k$) and the subscript $k$ from the RKHS $\calH_k$, whenever $k$ is the kernel on $\calX$ chosen by the database curator.

\subsection{Algorithm Template}
\label{sec:framework:template}

We propose the following general algorithm template for differentially private database release:
\begin{enumerate}
	\item Construct a consistent estimator $\hat{\mu}_X$ of the KME $\mu_X$ of $X$ using the private database.
	\item Obtain a perturbed version $\tilde{\mu}_X$ of the constructed estimate $\hat{\mu}_X$ to ensure differential privacy.
	\item Release a (potentially approximate) representation of $\tilde{\mu}_X$ in terms of a (possibly weighted) dataset $\{ (z_1, w_1), \ldots, (z_M, w_M) \} \subseteq \calX \times \IR$.
\end{enumerate}
The released representation should be such that $\sum_{m = 1}^M w_m k(z_m, \cdot)$ is a consistent estimator of the true KME $\mu_X$, i.e.~such that the RKHS distance between the two converges to $0$ in probability as the private database size $N$, and together with it the synthetic database size $M$, go to infinity.

Each step of this template admits several possibilities. For the first step we have discussed the standard empirical KME $\frac{1}{N} \sum_{n = 1}^N k(x_n, \cdot)$ with $x_1, \ldots, x_N$ i.i.d.~observations of $X$, but the framework remains valid with improved estimators such as \emph{kernel-based quadrature}~\cite{chen_herding_2010} or the \emph{shrinkage} kernel mean estimators of~\cite{muandet_kernel_2016}.

As the KMEs $\hat{\mu}_X$ and $\mu_X$ live in the RKHS $\calH$ of the kernel $k$, a natural mechanism for privatising $\hat{\mu}_X$ in the second step would be to follow~\cite{hall_differential_2013} and pointwise add to $\hat{\mu}_X$ a suitably scaled sample path $g$ of a Gaussian process with covariance function $k(\cdot, \cdot)$. This does ensure $(\varepsilon, \delta)$-differential privacy of the resulting function $\tilde{\mu}_X = \hat{\mu}_X + g$, but unfortunately $\tilde{\mu}_X \not\in \calH$, because the RKHS norm $\| g \|_{\calH}$ of a Gaussian process sample path with the same kernel $k$ is infinite almost surely~\cite{rasmussen_gaussian_2005}. While our framework allows pursuing this direction by, for example, moving to a larger function space that does contain the Gaussian process sample path, in this work we will present algorithms that achieve differential privacy by mapping $\hat{\mu}_X$ into a finite-dimensional Hilbert space and then employing the standard Laplace or Gaussian mechanisms to the finite coordinate vector.

Differential privacy is preserved under post-processing, but the third step does require some care to ensure that private data is not leaked. Specifically, when several possible (approximate) representations $\tilde{\mu}_X \approx \sum_{m = 1}^M w_m k(z_m, \cdot)$ in terms of a weighted dataset $(w_1, z_1), \ldots, (w_M, z_M)$ are possible, committing to a particular one reveals more information than just the function $\tilde{\mu}_X$ (consider, for example, the extreme case where the representation would be in terms of the private points $x_1, \ldots, x_N$). One thus needs to either control the privacy leak due to choosing a representation in a way that depends on the private data, or, as we do in our concrete algorithms below, choose a representation independently of the private data (but still minimising its RKHS distance to the privacy-protected $\tilde{\mu}_X$).

\subsection{Versatility}

Algorithms in our framework release a possibly weighted synthetic dataset $\{ (z_1, w_1), \ldots, (z_M, w_M) \} \subseteq \calX \times \IR$ such that $\sum_{m = 1}^M w_m k(z_m, \cdot)$ is a consistent estimator of the true KME $\mu_X$ of the data generating random variable $X$. This allows third-parties to perform a wide spectrum of statistical computation, all without having to worry about violating differential privacy:
\begin{enumerate}
	\item \emph{Kernel probabilistic programming}~\cite{scholkopf_computing_2015}: The versatility of our approach is greatly expanded thanks to the result of~\cite{scibior_consistent_2016}, who showed that under technical conditions, applying a continuous function $f$ to all points $z_m$ in the synthetic dataset yields a consistent estimator $\sum_{m = 1}^M w_m k_f(f(z_m), \cdot)$ of the KME $\mu_{f(X)}$ of the transformed random variable $f(X)$, even when the points $z_1, \ldots, z_M$ are not i.i.d. (as they may not be, depending on the particular synthetic database release algorithm).
	\item \emph{Consistent estimation of population statistics}: For any RKHS function $h \in \calH$, we have $\langle \mu_X, h \rangle_{\calH} = \IE[h(X)]$, so a consistent estimator of $\mu_X$ yields a consistent estimator of the expectation of $h(X)$. It can be evaluated using the reproducing kernel property:
	\begin{align}
	\IE[h(X)]
	&=
	\langle \mu_X, h \rangle_{\calH}
	\approx
	\left\langle \sum_{m = 1}^M w_m k(z_m, \cdot), h \right\rangle_{\calH}
	\nonumber
	\\&=
	\sum_{m = 1}^M w_m h(z_m)
	.
	\end{align}
	For example, approximating the indicator function $\mathds{1}_S$ of a set $S \subseteq \calX$ with functions in the RKHS allows estimating probabilities: $\IE[ \mathds{1}_S(X) ] = \IP[X \in S]$ (note that $\mathds{1}_S$ itself may not be an element of the RKHS).
	\item \emph{MMD estimation and two-sample testing}~\cite{gretton_kernel_2012}: Given another random variable $Y$ on $\calX$, one can consistently estimate the Maximum Mean Discrepancy (MMD) distance $\| \mu_X - \mu_Y \|_{\calH}$ between the distributions of $X$ and $Y$, and in particular to construct a two-sample test based on this distance. Given a sample $y_1, \ldots, y_L \sim Y$:
	\begin{equation*}
	\left\| \mu_X - \mu_Y \right\|_{\calH}
	\approx
	\left\| \sum_{m = 1}^M w_m k(z_m, \cdot) - \frac{1}{L} \sum_{l = 1}^L k(y_l, \cdot) \right\|_{\calH}
	,
	\end{equation*}
	which can again be evaluated using the reproducing property.
	\item \emph{Subsequent use of synthetic data}: Since the output of the algorithm is a (possibly weighted) database, third-parties are free to use this data for arbitrary purposes, such as training any machine learning model on this data. Models trained purely on this data can be released with differential privacy guaranteed; however, the accuracy of such models on real data remains an empirical question that is beyond the scope of this work.
\end{enumerate}

An orthogonal spectrum of versatility arises from the fact that the third step in the algorithm template can constrain the released dataset $(z_1, w_1), \ldots, (z_M, w_M)$ to be more convenient or more computationally efficient for further processing. For example, one could fix the weights to uniform $w_m = \frac{1}{M}$ to obtain an unweighted dataset, or to replace an expensive data type with a cheaper subset, such as requesting floats instead of doubles in the $z_m$'s. All this can be performed while an RKHS distance is available to control accuracy between $\tilde{\mu}_X$ and its released representation.

\subsection{Concrete Algorithms}
\label{sec:framework:concrete}

As a first illustrative example, we describe how a particular case of an existing, but inefficient synthetic database algorithm already fits into our framework. The \emph{exponential mechanism}~\citep{mcsherry_mechanism_2007} is a general mechanism for ensuring $\varepsilon$-differential privacy, and in our setting it operates as follows: given a similarity measure $s : \calX^N \times \calX^M \to \IR$ between (private) databases of size $N$ and (synthetic) databases of size $M$, output a random (synthetic) database $R$ with probability proportional to $\exp( \frac{\varepsilon}{2 \Delta_1} s(D, R))$, where $D$ is the actual private database and $\Delta_1$ is the $L_1$ sensitivity of $s$ w.r.t.~$D$. This ensures $\varepsilon$-differential privacy~\cite{mcsherry_mechanism_2007}.

To fit this into our framework, we can take $s(D, R) = - \| \mu_D - \mu_R \|_{\calH}$ to be the negative RKHS distance between the KMEs computed using $D$ and $R$, and achieve $\varepsilon$-differential privacy by releasing $R$ with probability proportional to $\exp( - \frac{\varepsilon}{2 \Delta_1} \| \mu_D - \mu_R \|_{\calH} )$. This solves steps 2 and 3 of our general algorithm template simultaneously, as it directly samples a concrete representation of a ``perturbed" KME $\mu_R$. The algorithm essentially corresponds to the SmallDB algorithm of~\citet{blum_learning_2008}, except for choosing the RKHS distance as a well-studied similarity measure between two databases.

The principal issue with this algorithm is its computational infeasibility except in trivial cases, as it requires sampling from a probability distribution supported on all potential synthetic databases, and employing an approximate sampling scheme can break the differential privacy guarantee of the exponential mechanism. In Sections~\ref{sec:perturb_in_synthetic_subspace} and~\ref{sec:perturb_in_random_features_RKHS} respectively, we describe two concrete synthetic database release algorithms that may possess failure modes where they become inefficient, but employing approximations in those cases can only affect their statistical accuracy, not the promise of differential privacy.

\begin{algorithm*}
	\caption{Differentially private database release via a synthetic data subspace of the RKHS}
	\label{alg:synthetic_subspace}
	\begin{algorithmic}[1]
		\REQUIRE database $D = \{ x_1, \ldots, x_N \} \subseteq \calX$, kernel $k$ on $\calX$, privacy parameters $\varepsilon > 0$ and $\delta > 0$
		\ENSURE $(\varepsilon, \delta)$-differentially private, weighted synthetic database (representing an estimate of $\mu_X$ in the RKHS $\calH$ of $k$)
		
		\STATE $M \gets M(N) \in \omega(1) \cap o(N^2)$, number of synthetic data points to use
		\STATE $z_1, \ldots, z_M \gets$ initialised deterministically or randomly from some distribution $q$ on $\calX$
		\STATE $\calH_M \gets \operatorname{Span}(\{ k(z_1, \cdot), \ldots, k(z_M, \cdot) \}) \leq \calH$
		\STATE $b_1, \ldots, b_F \gets$ orthonormal basis of $\calH_M$ (obtained using, e.g.~Gram-Schmidt)
		\STATE $\hat{\mu}_X \gets \frac{1}{N} \sum_{n = 1}^N k(x_n, \cdot)$, empirical KME of $X$ in $\calH$
		\STATE $\overline{\mu}_X \gets \sum_{f = 1}^F \langle b_f, \hat{\mu}_X \rangle_{\calH} b_f =: \sum_{f = 1}^F \alpha_f b_f$, projection of $\hat{\mu}_X$ onto $\calH_M$
		\STATE $\boldsymbol{\beta} \gets \boldsymbol{\alpha} + \calN(0, \frac{8 \ln(1.25/\delta)}{N^2 \varepsilon^2} I_F)$, an $(\varepsilon, \delta)$-differentially private version of the coordinate vector $\boldsymbol{\alpha}$ (Gaussian mechanism)
		\STATE $\tilde{\mu}_X \gets \sum_{f = 1}^F \beta_f b_f = \sum_{m = 1}^M w_m k(z_m, \cdot)$, re-expressed in terms of $k(z_m, \cdot)$'s \label{alg:synthetic_subspace:reexpress}
		\STATE \textbf{return} $(z_1, w_1), \ldots, (z_M, w_M)$
	\end{algorithmic}
\end{algorithm*}

\section{Perturbation in Synthetic-Data Subspace}
\label{sec:perturb_in_synthetic_subspace}

In this section we describe an instantiation of the framework proposed in Section~\ref{sec:framework} that achieves differential privacy of the KME by projecting it onto a finite-dimensional subspace of the RKHS spanned by feature maps $k(z_m, \cdot)$ of synthetic data points $z_1, \ldots, z_M$, and perturbing the resulting finite coordinate vector. To ensure differential privacy, the synthetic data points are chosen independently of the private database. As a result, statistical efficiency of this approach will depend on the choice of synthetic data points, with efficiency increasing if there are enough synthetic data points to capture the patterns in the private data. Therefore this algorithm is especially suited to the scenario discussed in Section~\ref{sec:introduction}, where a part of the database (or of a similar one) has already been published, as this can serve as a good starting set for the synthetic data points.

The setting where some observations from $X$ have already been released highlights the fact that differential privacy only protects against \emph{additional} privacy violation due to an individual deciding to contribute to the private database; if a particular user's data has already been published, differential privacy does not protect against privacy violations based on exploiting this previously published data.

The algorithm is formalised as Algorithm~\ref{alg:synthetic_subspace} above. Lines 1-2 choose synthetic data points $z_1, \ldots, z_M$ independently of the private data (only using the database size $N$). Lines 3-4 construct the linear subspace $\calH_M$ of $\calH$ spanned by feature maps of the chosen synthetic data points, and compute a (finite) basis for it. Only then the private data is accessed: the empirical KME $\hat{\mu}_X$ is computed (line 5), projected onto the subspace $\calH_M$ and expressed in terms of the precomputed basis (line 6). The basis coefficients of the projection are then perturbed to achieve differential privacy (line 7), and the perturbed element $\tilde{\mu}_X \in \calH_M$ is then re-expressed in terms of the spanning set containing feature maps of synthetic data points (line 8). This expansion is finally released to the public (line 9).

Line 1 stipulates that the number of synthetic data points $M \to \infty$ as $N \to \infty$, but asymptotically slower than $N^2$. This is to ensure that the privatisation noise added in the subspace $\calH_M$ to each coordinate is small enough overall to preserve consistency, as stated in the following Theorem~\ref{thm:synthetic_subspace_consistency}. This theorem assures us that Algorithm~\ref{alg:synthetic_subspace} produces a consistent estimator of the true KME $\mu_X$, if the synthetic data points are sampled from a distribution with sufficiently large support. Due to space constraints, all proofs appear in Appendix~\ref{sec:app:proofs}.

\begin{theorem}
	\label{thm:synthetic_subspace_consistency}
	Let $\calX$ be a compact metric space and $k$ a continuous kernel on $\calX$. If the synthetic data points $z_1, z_2, \ldots$ are sampled i.i.d.~from a distribution $q$ on $\calX$ such that the support of $X$ is included in the support of $q$, then Algorithm~\ref{alg:synthetic_subspace} outputs a consistent estimator of the KME $\mu_X$: $\sum_{m = 1}^M w_m k(z_m, \cdot) \stackrel{\IP}{\to} \mu_X$ as $N \to \infty$.
\end{theorem}

As discussed by~\citet{scibior_consistent_2016}, these assumptions are usually satisfied: $\calX$ can be taken to be compact whenever the data comes from measurements with any bounded range, and many kernels are continuous, including all kernels on discrete spaces (w.r.t.~to the discrete topology).

In order to use the output of Algorithm~\ref{alg:synthetic_subspace} in the very general \emph{kernel probabilistic programming} framework and obtain a consistent estimator of the KME $\mu_{f(X)}$ of $f(X)$ for \emph{any} continuous function $f$, there is a technical condition that the $L_1$ norm $\sum_{m = 1}^M |w_m|$ of the released weights may need to remain bounded by a constant as $N \to \infty$~\cite{scibior_consistent_2016}. This is not enforced by Algorithm~\ref{alg:synthetic_subspace}, but Theorem~\ref{thm:synthetic_subspace_regularization_consistency} in Appendix~\ref{app:sec:synthetic_subspace_consistency} shows how a simple regularisation in the final stage of the algorithm achieves this without breaking consistency (or privacy).

\begin{figure*}[!t]
	\centering
	\begin{subfigure}{.5\linewidth}
		\includegraphics[width=0.95\linewidth]{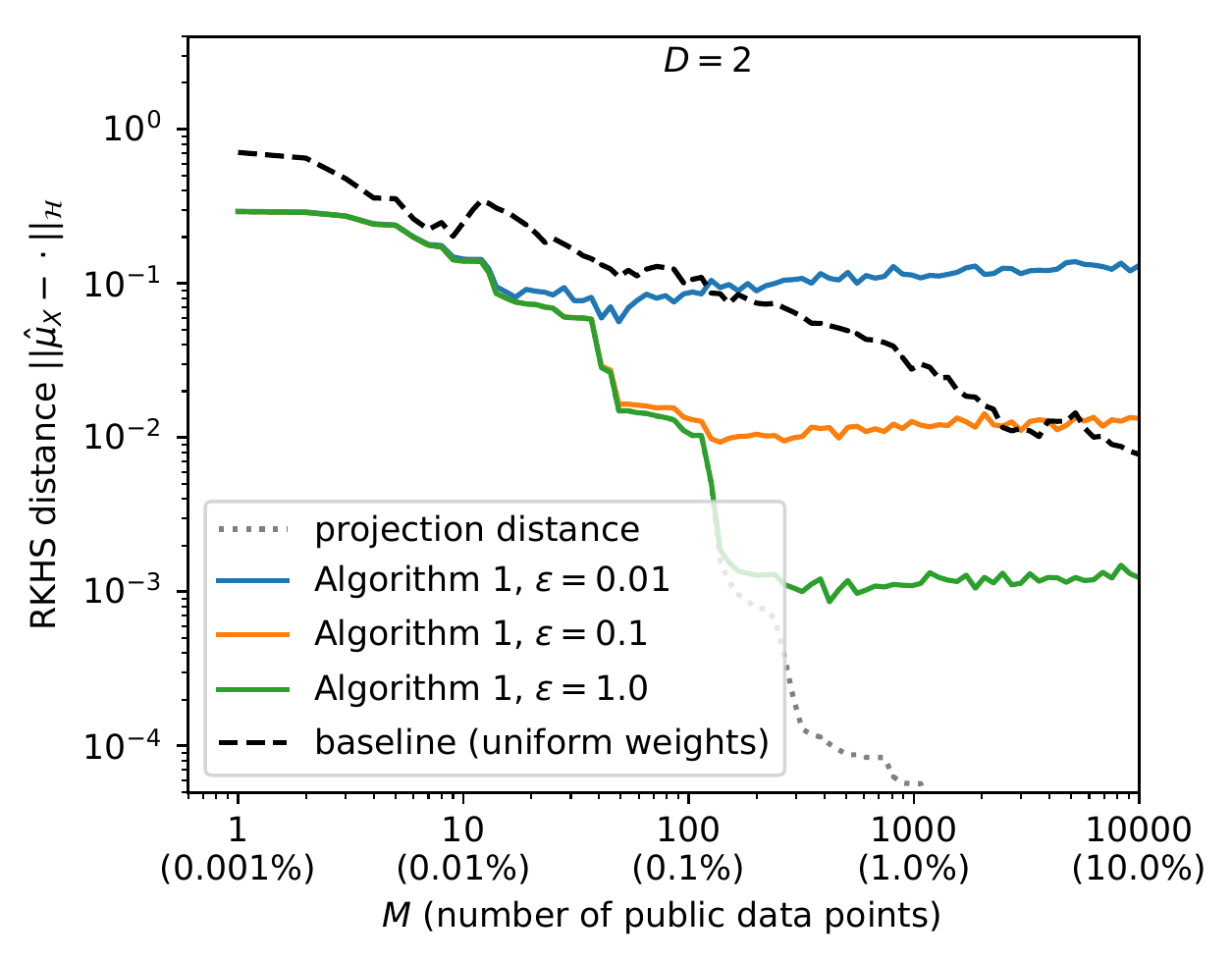}
	\end{subfigure}%
	\begin{subfigure}{.5\linewidth}
		\includegraphics[width=0.95\linewidth]{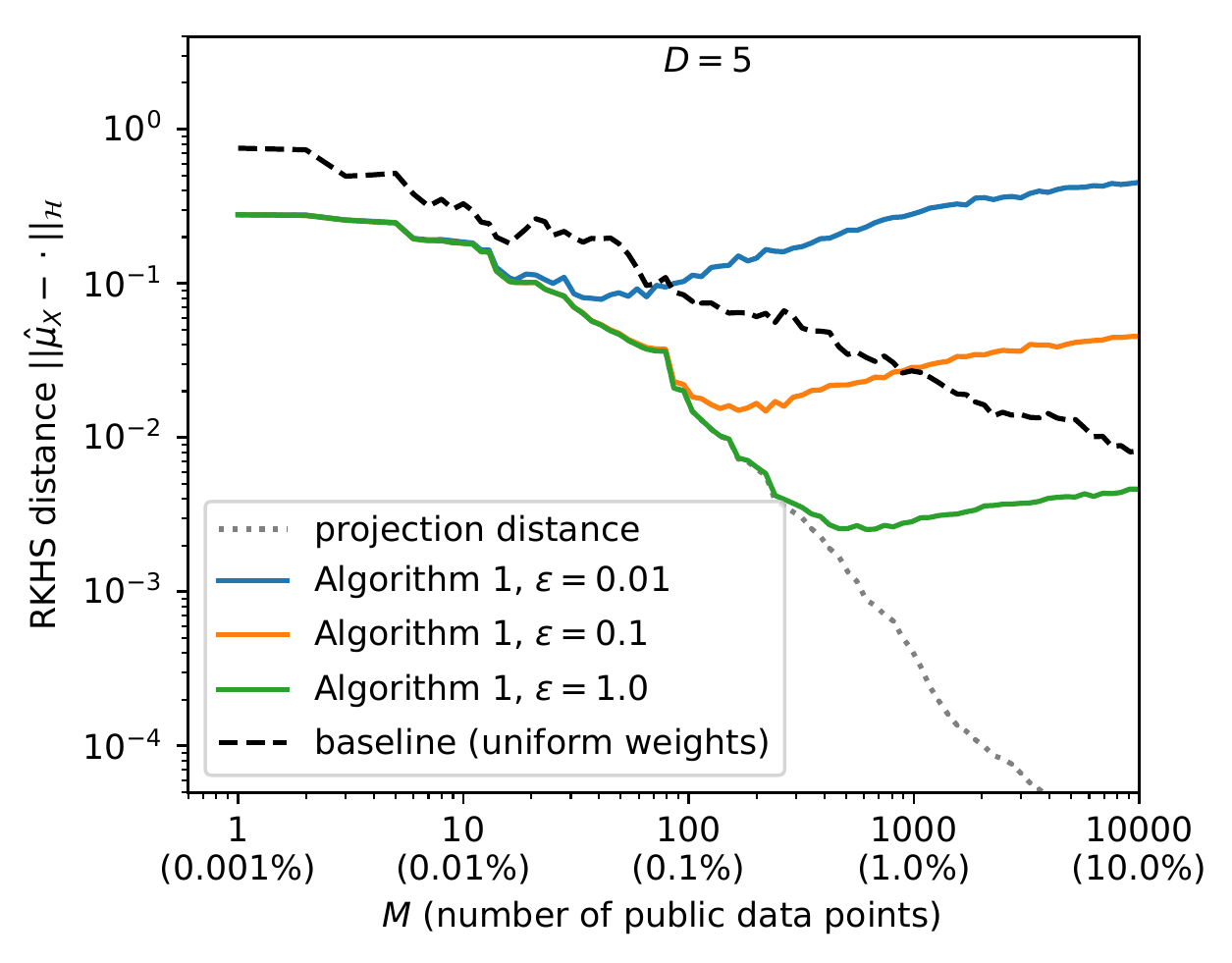}
	\end{subfigure}
	\vspace{-0.75em}
	\caption{\small{RKHS distance (lower is better) to the (private) empirical KME $\hat{\mu}_X$ computed using the entire private database of size $N = 100,000$. The dimension of the database was $D = 2$ (left) or $D = 5$ (right); please see Appendix~\ref{app:sec:experiments} for further details of the setup. Horizontally we varied $M$, the number of publicly releasable data points. Stricter privacy requirements (lower $\varepsilon$) naturally lead to lower accuracy. Increasing $M$ does not always necessarily improve accuracy, since a new public data point always increases the total amount of privatising noise that needs to be added, but this might not be outweighed by its positive contribution towards covering relevant parts of the input space. In all cases, for sufficiently small $M$ Algorithm~\ref{alg:synthetic_subspace} provided a significantly more accurate estimate than $\mu^{\text{baseline}}$.}}
	\label{fig:leaks}
\end{figure*}

The next result about Algorithm~\ref{alg:synthetic_subspace} shows that it is differentially private whenever $k(x, x) \leq 1$ for all $x \in \calX$. This is a weak assumption that holds for all normalised kernels, and can be achieved by simple rescaling for any bounded kernel (such that $\sup_{x \in \calX} k(x, x) < \infty$). When $\calX$ is a compact domain, all continuous kernels are bounded.

\begin{proposition}
	\label{prop:synthetic_subspace_privacy}
	If $k(x, x) \leq 1$ for all $x \in \calX$, then Algorithm~\ref{alg:synthetic_subspace} is $(\varepsilon, \delta)$-differentially private.
\end{proposition}

\begin{remark}
	\label{rem:privacy_params}
	One usually requires that $\delta$ decreases faster than polynomially with the database size $N$~\cite{dwork_algorithmic_2014}. The proof of Theorem~\ref{thm:synthetic_subspace_consistency} remains valid whenever $M(N) \in o(N^2 / \ln(1.25 / \delta(N)))$, so for example we could have $\delta(N) = e^{-\sqrt{N}}$ and $M(N) \in o(N^{3/2})$.
  \qed
\end{remark}

For a finite private database, actual performance will heavily depend on how the synthetic data points are chosen. We consider the following two extreme scenarios:
\begin{enumerate}
	\vspace{-0.5em}
	\item \emph{No publishable subset}: No rows of the private database are, or can be made public unmodified.
	\item \emph{Publishable subset}: A small proportion of the private database is already public, or can be made public.
\end{enumerate}

\begin{proposition}[Algorithm 1, No publishable subset]
\label{prop:alg1_rate_no_publishable_subset}
Say $\calX$ is a bounded subset of $\IR^D$, the kernel $k$ is Lipschitz, and the synthetic data points $z_1, z_2, \ldots$ are sampled i.i.d.~from a distribution $q$ with density bounded away from $0$ on any bounded subset of $\IR^D$. Then $M = M(N)$ can be chosen so that the output of Algorithm~\ref{alg:synthetic_subspace} converges to the true KME $\mu_X$ in RKHS norm at a rate $\calO_p(N^{-1/(D+1+c)})$, where $c$ is any fixed positive number $c > 0$.
\end{proposition}

Unsurprisingly, the convergence rate deteriorates with input dimension $D$, since without prior information about the private data manifold it is increasingly difficult for randomly sampled synthetic points to capture patterns in the private data. One of the main strengths of KMEs is that the empirical estimator converges to the true embedding at a rate $\calO_p(N^{-1/2})$ independently of the input dimension $D$, so we see that in this unfavourable scenario Algorithm~\ref{alg:synthetic_subspace} incurs a substantial privacy cost in high dimensions. On the other hand, if a small, but fixed proportion of the private database is publishable, then Algorithm~\ref{alg:synthetic_subspace} incurs no privacy cost in terms of the convergence rate:

\begin{proposition}[Algorithm 1, Publishable subset]
\label{prop:alg1_rate_publishable_subset}
Say that a fixed proportion $\eta$ of the private database can be published unmodified. Using this part of the database as the synthetic data points, Algorithm~\ref{alg:synthetic_subspace} outputs a consistent estimator of $\mu_X$ that converges in RKHS norm at a rate $\calO_p(N^{-1/2})$.
\end{proposition}

Note that in this scenario the rate $\calO_p(N^{-1/2})$ can be also achieved by uniform weighting of the synthetic data points, since $\hat{\mu}^{\text{baseline}} := \frac{1}{M} \sum_{m = 1}^M k(z_m, \cdot)$ with $z_m = x_m$ is already a consistent estimator of $\mu_X$ (although based on a much smaller sample size $M = \eta N \ll N$). The purpose of Algorithm~\ref{alg:synthetic_subspace} is to find (generally non-uniform) $w_1, \ldots, w_M$ that reweight the public data points using the information in the large private dataset, but respecting differential privacy. Proposition~\ref{prop:alg1_rate_publishable_subset} confirmed theoretically that this does not hurt the convergence rate, while Figure~\ref{fig:leaks} shows empirically on two synthetic datasets of dimensions $D = 2$ and $D = 5$ that Algorithm~\ref{alg:synthetic_subspace} can in fact yield more accurate estimates of the KME than $\hat{\mu}^{\text{baseline}}$, especially when the proportion of public data points is small. This is encouraging, since obtaining permission to publish a larger subset of the private data unchanged will usually come at an increased cost. The ability to instead reweight a smaller public dataset in a differentially private manner using Algorithm~\ref{alg:synthetic_subspace} is therefore useful.

\begin{algorithm*}
	\caption{Differentially private database release via a random features RKHS}
	\label{alg:random_features}
	\begin{algorithmic}[1]
		\REQUIRE database $D = \{ x_1, \ldots, x_N \} \subseteq \calX$, kernel $k$ on $\calX$, privacy parameters $\varepsilon > 0$ and $\delta > 0$
		\ENSURE $(\varepsilon, \delta)$-differentially private, weighted synthetic database (representing an estimate of $\mu_X$ in the RKHS $\calH$ of $k$)

		\STATE $J \gets J(N) \in \omega(1) \cap o(N^2)$, number of random features to use
		\STATE $\phi \gets$ random feature map $\calX \mapsto \IR^J$ for kernel $k$ with $J$ features
		\STATE $\hat{\mu}^{\phi}_X \gets \frac{1}{N} \sum_{n = 1}^N \phi(x_n) \in \IR^J$, empirical KME of $X$ in RKHS $\calH_{\phi}$ of the random features kernel $k_{\phi}(\cdot, \cdot) := \phi(\cdot)^T \phi(\cdot)$
		\STATE $\tilde{\mu}^{\phi}_X \gets \hat{\mu}^{\phi}_X + \calN(0, \frac{8 \ln(1.25 / \delta)}{N^2 \varepsilon^2} I_J)$, an $(\varepsilon, \delta)$-differentially private version of the vector $\hat{\mu}^{\phi}_X$ (Gaussian mechanism)
		\STATE $M \gets M(N) \geq N$, number of synthetic expansion points to use for representing $\tilde{\mu}^{\phi}_X$
		\STATE $(z_1, w_1), \ldots, (z_M, w_M) \gets$ approximate $\tilde{\mu}^{\phi}_X$ in the RKHS $\calH_{\phi}$ using a Reduced set method:
		\vspace{-0.5em}
		\begin{equation}
		(z_1, w_1), \ldots, (z_M, w_M)
		\approx
		\argmin_{\substack{(z'_1, w'_1), \ldots, (z'_M, w'_M)\text{ s.t. } \sum_m |w'_m| \leq 1}}
		\left\|
		\sum_{m = 1}^M w'_m \phi(z'_m) - \tilde{\mu}^{\phi}_X
		\right\|_{\calH_{\phi}}
		\label{eq:random_features_preimage_minimisation}
		\end{equation}
		\vspace{-1em}
		\STATE \textbf{return} $(z_1, w_1), \ldots, (z_M, w_M)$
	\end{algorithmic}
\end{algorithm*}

\section{Perturbation in Random-Features RKHS}
\label{sec:perturb_in_random_features_RKHS}

Another approach to ensuring differential privacy is to map the potentially infinite dimensional RKHS $\calH$ of $k$ into a different, finite-dimensional RKHS $\calH_{\phi}$ using random features~\cite{rahimi_random_2007}, privacy-protect the finite coordinate vector in this space~\cite{chaudhuri_differentially_2011}, and then employ a reduced set method to find an expansion of the resulting RKHS element in terms of synthetic data points. In contrast to Algorithm~\ref{alg:synthetic_subspace}, both the weights and locations of synthetic data points can be optimised here.

The algorithm is formalised as Algorithm~\ref{alg:random_features} above. Lines 1-2 pick the number $J = J(N)$ of random features to use, and construct a random feature map $\phi$ with that many features. Lines 3-4 compute the empirical KME of $X$ in the RKHS $\calH_{\phi}$ corresponding to the kernel induced by the random features, and then privacy-protect the resulting finite, real-valued vector. Lines 5-6 run a blindly initialised Reduced set method to find a weighted synthetic dataset whose KME in $\calH_{\phi}$ is close to the privacy-protected KME of the private database. Line 7 releases this weighted dataset to the public.

The following theorem confirms that Algorithm~\ref{alg:random_features} outputs a consistent estimator of the true KME $\mu_X$, provided that the optimisation problem (\ref{eq:random_features_preimage_minimisation}) is solved exactly, and the random features converge to the kernel $k$ uniformly on $\calX$. On compact sets $\calX$ this requirement is satisfied by general schemes such as \emph{random Fourier features} and \emph{random binning} for shift-invariant kernels~\citep{rahimi_random_2007}, or by random features for dot product kernels~\cite{kar_dot_2012}.

\begin{theorem}
	\label{thm:random_features_consistency}
	If $\phi(\cdot)^T \phi(\cdot) \to k(\cdot, \cdot)$ converges uniformly in $\calX \times \calX$ as $J \to \infty$, then the output of Algorithm~\ref{alg:random_features} is a consistent estimator of the true KME $\mu_X$ as $N \to \infty$.
\end{theorem}

Moreover, a uniform convergence rate for the random features, such as the one for random Fourier features by~\citet{sriperumbudur_optimal_2015}, can be used to derive a convergence rate for the output of Algorithm~\ref{alg:random_features}:

\begin{proposition}
If $\phi(\cdot)^T \phi(\cdot) \to k(\cdot, \cdot)$ converges uniformly in $\calX \times \calX$ at a rate $\calO_p(J^{-1/2})$ as $J \to \infty$, then $J = J(N)$ can be chosen so that the output of Algorithm~\ref{alg:random_features} converges to the true KME $\mu_X$ at a rate $\calO_p(N^{-1/3})$.
\end{proposition}

\begin{figure*}[!t]
	\centering
	\begin{subfigure}{.5\linewidth}
		\includegraphics[width=0.92\linewidth]{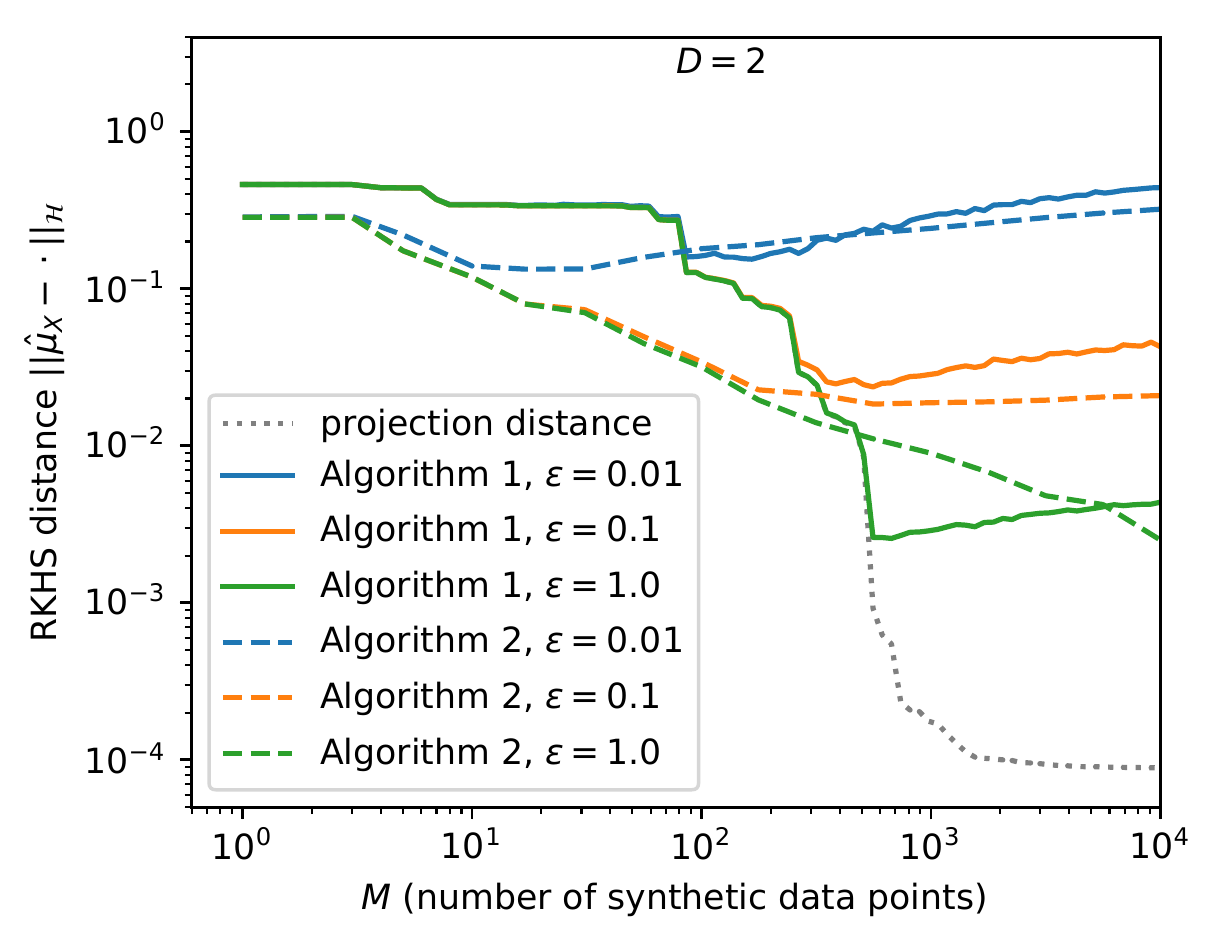}
	\end{subfigure}%
	\begin{subfigure}{.5\linewidth}
		\includegraphics[width=0.92\linewidth]{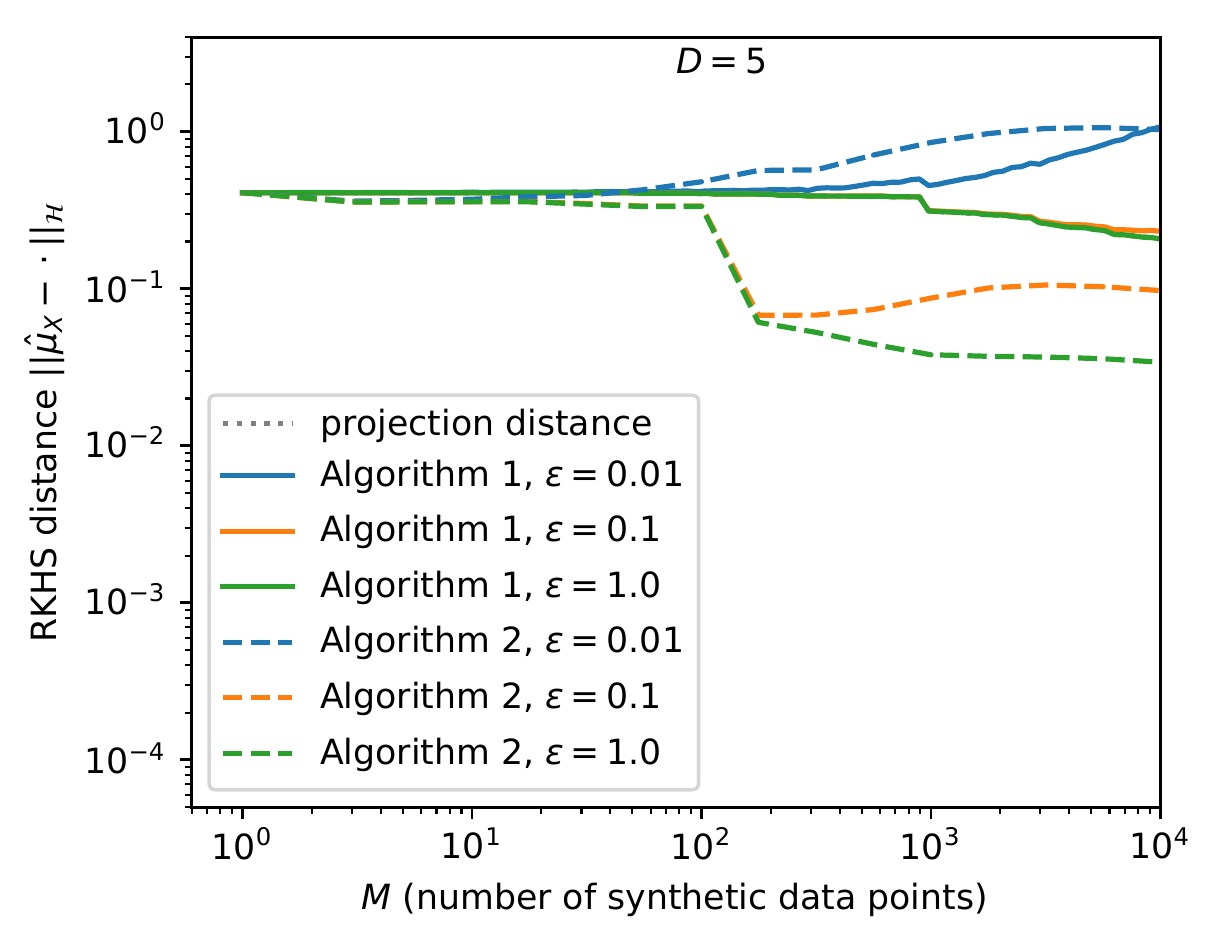}
	\end{subfigure}
	\vspace{-0.75em}
	\caption{\small{RKHS distance (lower is better) to the (private) empirical KME $\hat{\mu}_X$ computed using the same databases as in Figure~\ref{fig:leaks}, of dimensions $D = 2$ (left) and $D = 5$ (right), but this time without a publishable subset. The synthetic data points for Algorithm 1 were therefore sampled from a wide Gaussian distribution; please see Appendix~\ref{app:sec:experiments} for further details. Algorithm~\ref{alg:random_features} is capable of outperforming Algorithm~\ref{alg:synthetic_subspace} thanks to its ability to optimise the synthetic data point locations, but this depends on the precise optimisation procedure used and the optimisation problem becomes harder in higher dimensions.}}
	\label{fig:nodata}
\end{figure*}

The empirical KME of the private database $\hat{\mu}_X$ converges at a rate $\calO_p(N^{-1/2})$, so we see that under perfect optimisation, the privacy cost incurred by Algorithm~\ref{alg:random_features} is a factor of $N^{1/6}$. In practice performance will also depend on the Reduced set method used, and the computational budget allocated to it. Figure~\ref{fig:nodata} shows how the incurred error (in terms of RKHS distance) varies with the number of synthetic data points $M$. The additional ability of Algorithm~\ref{alg:random_features} to optimise the \emph{locations} of the synthetic data points (rather than just the weights, as in Algorithm~\ref{alg:synthetic_subspace}) seems to be more helpful in the higher-dimensional case $D = 5$, where the randomly sampled synthetic data points are less likely to land close to private data points.

\begin{proposition}
	\label{prop:random_features_privacy}
	If $\| \phi(x) \|_2 \leq 1$ for all $x \in \calX$, then Algorithm~\ref{alg:random_features} is $(\varepsilon, \delta)$-differentially private.
\end{proposition}

This $L_2$-boundedness requirement on the random feature vectors $\phi(x)$ is reasonable under the weak assumption $k(x, x) \leq 1$ for all $x \in \calX$ discussed in Section~\ref{sec:perturb_in_synthetic_subspace}, as in that case $\| \phi(x) \|_2^2 = \phi(x)^T \phi(x) \approx k(x, x) \leq 1$.

\section{Related Work}
\label{sec:related_work}

Synthetic database release algorithms with a differential privacy guarantee have been studied in the literature before. \citet{machanavajjhala2008privacy} analyzed such a procedure for count data, ensuring privacy by sampling a distribution and then synthetic counts from a Dirichlet-Multinomial posterior. \citet{blum_learning_2008} studied the exponential mechanism applied to synthetic database generation, which leads to a very general, but unfortunately inefficient algorithm (see also Section~\ref{sec:framework:concrete}). \citet{wasserman_statistical_2008} provided a theoretical comparison of this algorithm to sampling synthetic databases from deterministically smoothed, or randomly perturbed histograms. Unlike our approach, these algorithms achieve differential privacy by sampling synthetic data points from a specific distribution, where resorting to approximate sampling can break the privacy guarantee. In our framework we propose to arrive at the synthetic database using a reduced set method, where poor performance could affect statistical usefulness of the synthetic database, but cannot break its differential privacy.

\citet{zhou_differential_2009} and \citet{kenthapadi_privacy_2012} proposed randomised database compression schemes that yield synthetic databases useful for particular types of algorithms, while guaranteeing differential privacy. The former compresses the number of data points using a random linear or affine transformation of the entire database, and the result can be used by procedures that rely on the empirical covariance of the original data. The latter compresses the number of data point dimensions while approximately preserving distances between original, private data points.

Differentially private learning in a RKHS has also been studied, with~\citet{chaudhuri_differentially_2011} and \citet{rubinstein_svm_2012} having independently presented release mechanisms for the result of an empirical risk minimisation procedure (such as a SVM). Similarly to our Algorithm~\ref{alg:random_features}, they map data points into a finite-dimensional space defined by random features and carry out the privacy-protecting perturbation there. However, they do not require the final stage of invoking a Reduced set method to construct a synthetic database, because the output (such as a trained SVM) is only used for evaluation on test points, for which it suffices to additionally release the used random feature map $\phi$.

As our framework stipulates privacy-protecting an empirical KME, which is a function $\calX \to \IR$, the work on differential privacy for functional data is of relevance. \citet{hall_differential_2013} showed how an RKHS element can be made differentially private via pointwise addition of a Gaussian process sample path, but as discussed in Section~\ref{sec:framework:template}, the resulting function is no longer an element of the RKHS. Recently,~\citet{alda_bernstein_2017} proposed a general Bernstein mechanism for $\varepsilon$-differentially private function release. The released function can be evaluated pointwise arbitrarily many times, but again, the geometry of the RKHS to which the unperturbed function belonged cannot be easily exploited anymore.

\section{Discussion}
\label{sec:discussion}

We proposed a framework for constructing differentially private synthetic database release algorithms, based on the idea of using KMEs in RKHS as intermediate database representations.
To justify our framework, we presented two concrete algorithms and proved theoretical results guaranteeing their consistency and differential privacy.
We also studied their finite-sample convergence rates, and provided empirical illustrations of their performance on synthetic datasets.
We believe that exploring other instantiations of this framework, and comparing them theoretically and empirically, can be a fruitful direction for future research.

The i.i.d.~assumption on database rows can be relaxed. For example, if they are identically distributed (as a random variable $X$), but not necessarily independent, the framework remains valid as long as a consistent estimator of the KME $\mu_X$ can be constructed from the database rows. A common situation where this arises is, for example, duplication of database rows due to user error.

\section*{Acknowledgements}
The authors would like to thank Bharath Sriperumbudur and the anonymous reviewers for helpful feedback.

\begin{small}
\bibliography{RKHSprivacy}
\bibliographystyle{icml2018}
\end{small}

\clearpage
\newpage
\onecolumn
\appendix

\section*{APPENDIX: Differentially Private Database Release via Kernel Mean Embeddings}

\section{Proofs}
\label{sec:app:proofs}

Here we provide proofs for the results stated in the main text, together with additional supporting lemmas required for these proofs.

\subsection{Algorithm 1 (Synthetic Data Subspace): Consistency}
\label{app:sec:synthetic_subspace_consistency}

Before proving Theorem 2, we obtain a Lemma showing that the ``projection error" incurred due to projecting the KME $\hat{\mu}_X$ onto the finite-dimensional subspace $\calH_M$ spanned by the synthetic data points, quantified by the RKHS distance between $\hat{\mu}_X$ and the projection $\overline{\mu}_X$, converges to $0$ as $N \to \infty$:

\begin{lemma}
	\label{lem:projection_consistency}
	Let $\calX$ be a compact metric space and $k : \calX \times \calX \to \IR$ a continuous kernel on $\calX$. Suppose that the synthetic data points $z_1, z_2, \ldots$ are sampled i.i.d.~from a probability distribution $q$ on $\calX$. If the support $\supp(X)$ of $X$ is included in the support of $q$, then
	\begin{equation}
	\left\| \overline{\mu}_X - \hat{\mu}_X \right\|_{\calH}
	\stackrel{\IP}{\to}
	0
	\text{ as }
	N \to \infty
	.
	\end{equation}
	\begin{proof}
		Let $\varepsilon > 0$. As $k$ is continuous on $\calX \times \calX$, which as a product of compact spaces is itself compact by Tychonoff's theorem, the kernel $k$ is uniformly continuous and in particular there exists $\delta > 0$ such that for all $x, x' \in \calX$ we have $| k(x, x) - k(x, x') | < \varepsilon^2 / 2$ whenever $\| x - x' \|_{\calX} < \delta$. As $\calX$ is compact, it is totally bounded, and thus so is its subset $\supp(X)$. Therefore $\supp(X)$ can be covered with finitely many open balls $B_1, \ldots, B_{K}$ of radius $\delta / 2$. Let the sequence $z_1, z_2, \ldots$ be sampled i.i.d.~from $q$, and let $E_M$ be the event that at least ones of these $K$ balls contains no element of $z_1, \ldots, z_M$. Since $\supp(X) \subseteq \supp(q)$ by assumption, we have $q(B_k) > 0$ for all $k = 1, \ldots, K$ and therefore $\IP[ E_M ] \to 0$ as $M \to \infty$.
		
		Note that if all $K$ balls contain an element of $z_1, \ldots, z_M$ (i.e., $E_M^C$ holds), then for each $x \in \supp(X)$ one can find $1 \leq m(x) \leq M$ such that $\| x - z_{m(x)} \| < \delta/2 + \delta/2 = \delta$. In that case
		\begin{align}
		\left\| \overline{\mu}_X - \hat{\mu}_X \right\|_{\calH}
		&=
		\inf_{h \in \calH_M} \left\| h - \hat{\mu}_X \right\|_{\calH}
		& \text{[property of projection]}\nonumber
		\\&\leq
		\left\| \frac{1}{N} \sum_{n = 1}^N k(z_{m(x_n)}, \cdot) - \hat{\mu}_X \right\|_{\calH}
		& \text{[as $\textstyle\frac{1}{N} \textstyle\sum_{n = 1}^N k(z_{m(x_n)}, \cdot) \in \calH_M$]}\nonumber
		\\&\leq
		\frac{1}{N} \sum_{n = 1}^N \left\| k(z_{m(x_n)}, \cdot) - k(x_n, \cdot) \right\|_{\calH}
		& \text{[Triangle inequality]}\nonumber
		\\&<
		\frac{1}{N} \sum_{n = 1}^N \varepsilon
		& \text{[see below]}\nonumber
		\\&=
		\varepsilon
		,
		\end{align}
		where we have used the reproducing property, the Triangle inequality and our choices of $\delta$ and $z_{m(x_n)}$ to see that for all $1 \leq n \leq N$,
		\begin{align}
		\left\| k(z_{m(x_n)}, \cdot) - k(x_n, \cdot) \right\|_{\calH}
		&=
		\langle k(z_{m(x_n), \cdot} - k(x_n, \cdot), k(z_{m(x_n), \cdot} - k(x_n, \cdot) \rangle_{\calH}^{1/2}
		\\&=
		\Big( k(z_{m(x_n)}, z_{m(x_n)})
		- 2 k(z_{m(x_n)}, x_n)
		+ k(x_n, x_n) \Big)^{1/2}
		\\&\leq
		\big( | k(z_{m(x_n)}, z_{m(x_n)}) - k(z_{m(x_n)}, x_n) |
		+ | k(x_n, x_n) - k(z_{m(x_n)}, x_n) | \big)^{1/2}
		\\&<
		\left( \frac{\varepsilon^2}{2} + \frac{\varepsilon^2}{2} \right)^{1/2}
		= \varepsilon
		.
		\end{align}

		Hence we have that $\IP\left[
		\left\| \overline{\mu}_X - \hat{\mu}_X \right\|_{\calH}
		> \varepsilon
		\right]
		\leq \IP[ E_M ]
		\to 0
		\text{ as }
		M \to \infty$. But since $\varepsilon > 0$ was arbitrary and $M \to \infty$ as $N \to \infty$ by construction, the claimed convergence in probability result follows from definition.
	\end{proof}
\end{lemma}

\begin{hthm}[\ref{thm:synthetic_subspace_consistency}.]
	\label{app:thm:synthetic_data_consistency}
	Let $\calX$ be a compact metric space and $k : \calX \times \calX \to \IR$ a continuous kernel on $\calX$. Suppose that the synthetic data points $z_1, z_2, \ldots$ are sampled i.i.d.~from a probability distribution $q$ on $\calX$. If the support of $X$ is included in the support of $q$, then Algorithm~\ref{alg:synthetic_subspace} outputs a consistent estimator of the kernel mean embedding $\mu_X$ in the sense that
	\begin{equation}
	\sum_{m = 1}^M w_m k(z_m, \cdot)
	\stackrel{\IP}{\to}
	\mu_X
	\hspace{2em} \text{as } N \to \infty
	.
	\end{equation}
	\begin{proof}
		Using the Triangle inequality, we can upper bound the RKHS distance between the output $\tilde{\mu}_X$ of Algorithm~\ref{alg:synthetic_subspace} and the true kernel mean embedding $\mu_X$ as follows:
		\begin{equation}
		\left\| \tilde{\mu}_X - \mu_X \right\|_{\calH}
		\leq
		\underbrace{
			\left\| \tilde{\mu}_X - \overline{\mu}_X \right\|_{\calH}
		}_{\text{privacy error}}
		+
		\underbrace{
			\left\| \overline{\mu}_X - \hat{\mu}_X \right\|_{\calH}
		}_{\text{projection error}}
		+
		\underbrace{
			\left\| \hat{\mu}_X - \mu_X \right\|_{\calH}
		}_{\text{finite sample error}}
		.
		\label{eq:synthetic_subspace_consistency:triangle_bound}
		\end{equation}
		The finite sample error tends to $0$ as $N \to \infty$ by the law of large numbers, while the projection error tends to $0$ as $N \to \infty$ by Lemma~\ref{lem:projection_consistency}. For the privacy error, using orthonormality of the basis $
		\{ b_1, \ldots, b_F \}$ we have
		\begin{equation}
		\left\| \tilde{\mu}_X - \overline{\mu}_X \right\|_{\calH}^2
		=
		\left\| \sum_{f = 1}^F (\beta_f - \alpha_f) b_f \right\|_{\calH}^2
		=
		\sum_{f = 1}^F (\beta_f - \alpha_f)^2
		=
		\frac{8 \ln(1.25 / \delta)}{N^2 \varepsilon^2} F \frac{1}{F} \sum_{f = 1}^F \calN(0, 1)^2
		.
		\end{equation}
		As a function of $N$, the size of the basis $F \in \IN$ is a non-decreasing function, so it either converges to some $L \in \IN$, in which case the obtained expression clearly tends to $0$ as $N \to \infty$ with probability $1$, or $F \to \infty$ as $N \to \infty$. In this latter case $\frac{1}{F} \sum_{f = 1}^F \calN(0, 1)^2 \to 1$ as $N \to \infty$ a.s.~by the strong law of large numbers, and $F / N^2 \to 0$ as $N \to \infty$ since $F \leq M = o(N^2)$. Hence the privacy error goes to $0$ as $N \to \infty$ either way, as required to complete the proof.
	\end{proof}
\end{hthm}

\begin{theorem}
	\label{thm:synthetic_subspace_regularization_consistency}
	Suppose that the kernel $k$ is $c_0$-universal~\cite{sriperumbudur_universality_2011} and $f$ is any continuous function mapping from $\calX$ to some space $\mathcal{Y}$. Let $C \geq 1$ be any finite constant. If line
	7
	of Algorithm~\ref{alg:synthetic_subspace} is replaced with a regularised reduced set method solving the constrained minimisation problem
	\begin{equation}
	\mathbf{w}
	=
	\argmin_{\mathbf{u}: \| \mathbf{u} \|_1 \leq C}
	\left\|
	\tilde{\mu}_X
	-
	\sum_{m = 1}^M u_m k(z_m, \cdot)
	\right\|_{\calH}
	,
	\label{eq:synthetic_subspace_regularized_minimization}
	\end{equation}
	then the points output by Algorithm~\ref{alg:synthetic_subspace} yield a consistent estimator of the kernel mean embedding $\IE[k(f(X), \cdot)]$ of $f(X)$ in the sense that
	\begin{equation}
	\sum_{m = 1}^M w_m k(f(z_m), \cdot)
	\stackrel{\IP}{\to}
	\mu_{f(X)}
	\hspace{2em} \text{as } N \to \infty
	.
	\end{equation}
	\begin{proof}
		Let ${\mu}_X^{\text{out}} := \sum_{m = 1}^M w_m k(z_m, \cdot)$ be the RKHS element output by Algorithm~\ref{alg:synthetic_subspace} after adding the stated regularisation. First we show that despite the regularisation, ${\mu}_X^{\text{out}}$ remains a consistent estimator of the true kernel mean embedding $\mu_X$ as $N \to \infty$.
		
		The modification introduces an additional regularisation error term $\| {\mu}_X^{\text{out}} - \tilde{\mu}_X \|_{\calH}$ into the upper bound on $\| {\mu}_X^{\text{out}} - \mu_X \|$, compared to the corresponding bound (\ref{eq:synthetic_subspace_consistency:triangle_bound}) in the proof of Theorem~\ref{thm:synthetic_subspace_consistency}. So to show the first desired consistency result, it remains to show that this extra regularisation error term converges to $0$ in probability as $N \to \infty$. To this end, let $\varepsilon > 0$ be arbitrary. Define $\delta > 0$, the sequence $z_1, z_2, \ldots$ and $m(x)$ for $x \in \calX$ as in the proof of Lemma~\ref{lem:projection_consistency}. Note that the RKHS element $\frac{1}{N} \sum_{n = 1}^N k(z_{m(x_n)}, \cdot)$ is in the feasible set of the regularised minimisation problem (\ref{eq:synthetic_subspace_regularized_minimization}), because the sum of absolute values of expansions coefficients is
		\begin{equation}
		\sum_{m = 1}^M \sum_{n : m(x_n) = n} \frac{1}{M}
		=
		\sum_{n = 1}^N \frac{1}{N}
		=
		1
		\leq C
		\end{equation}
		Therefore the regularisation error can be upper bounded as
		\begin{align}
		\| {\mu}_X^{\text{out}} - \tilde{\mu}_X \|_{\calH}
		&\leq
		\left\|
		\frac{1}{N} \sum_{n = 1}^N k(z_{m(x_n)}, \cdot) - \tilde{\mu}_X
		\right\|_{\calH}
		& \text{[property of min]}\nonumber
		\\ &\leq
		\left\|
		\tilde{\mu}_X - \hat{\mu}_X
		\right\|_{\calH}
		+
		\left\|
		\hat{\mu}_X - \frac{1}{N} \sum_{n = 1}^N k(z_{m(x_n)}, \cdot)
		\right\|_{\calH}
		& \text{[Triangle inequality]}\nonumber
		\end{align}
		The first term goes to $0$ as $N \to \infty$ by the argument given in the proof of Theorem~\ref{thm:synthetic_subspace_consistency}. The probability that the second term is larger than $\varepsilon$ converges to $0$ as $N \to \infty$ using the argument given in the proof of Lemma~\ref{lem:projection_consistency}. Hence we have the desired convergence of the modified Algorithm~\ref{alg:synthetic_subspace}'s output ${\mu}_X^{\text{out}}$ to the true kernel mean embedding $\mu_X$ as $N \to \infty$, in probability.
		
		This means that the modified algorithm still outputs a consistent estimator of the kernel mean embedding of $\mu_X$. Moreover, the weights in the released finite expansion now have their $L_1$ norm $\sum_{m = 1}^M |w_m|$ bounded by the constant $C$ by construction, so Theorem 1 of~\cite{scibior_consistent_2016} applies and gives the desired conclusion regarding consistency of the estimator for the kernel mean embedding $\mu_{f(X)}$ of $f(X)$.
	\end{proof}
\end{theorem}

\subsection{Algorithm 1 (Synthetic Data Subspace): Convergence Rates}
\label{app:sec:synthetic_subspace_convergence_rates}

Towards proving the convergence rate of Proposition~\ref{prop:alg1_rate_no_publishable_subset}, we will make use of the following Lemma~\ref{lem:projection_convergence_rate}, which is a refinement of the corresponding consistency result of Lemma~\ref{lem:projection_consistency} above. It uses the Lipschitz assumption on the kernel to establish a quantitative dependence between $\varepsilon$ and $\delta$, and the condition on $q$ to establish a dependence between $\delta$, $K$ and $\IP[E_M]$.

\begin{lemma}
\label{lem:projection_convergence_rate}
Suppose that $\calX$ is a bounded subset of $\IR^D$, the kernel $k$ is Lipschitz with some Lipschitz constant $L \in \IR^{+}$, and the synthetic data points $z_1, z_2, \ldots$ are sampled i.i.d.~from a distribution $q$ whose density is bounded away from $0$ on any bounded subset of $\IR^D$. Then
\begin{equation*}
\forall{\gamma \in (0, 1), a > 0}
\hspace{1.5em}
\exists{C \in \IR, \varepsilon_0 > 0}
\hspace{1.5em}
\forall{\varepsilon \in (0, \varepsilon_0)}
\hspace{1.5em}
M \geq C \varepsilon^{-2D-a}
\hspace{0.75em} \Rightarrow \hspace{0.75em}
\IP\left[ \| \hat{\mu}_X - \bar{\mu}_X \|_{\calH} \geq \varepsilon \right]
\leq
\gamma
.
\end{equation*}
\begin{proof}
Let $\gamma \in (0, 1)$ and $a > 0$.
Suppose for the moment that $C$ and $\varepsilon_0$ have already been chosen based on $\calX, q, \gamma, a$ and based on the Lipschitz constant $L$ of the kernel $k$. Let $\varepsilon \in (0, \varepsilon_0)$ and suppose that $M \geq C \varepsilon^{-2D-a}$.

Define $\delta = \frac{\varepsilon^2}{2L}$ and let $B_1, \ldots, B_K$ be a covering of $\supp(X)$ with $K$ open balls of radii $\frac{\delta}{2}$. By the Lipschitz property
\begin{equation*}
\| x - x' \|_{\calX} < \delta
\hspace{0.75em}\Rightarrow\hspace{0.75em}
|k(x, x') - k(x, x)|
\leq L \| x - x' \|_{\calX}
< L \delta
= \frac{\varepsilon^2}{2}
\end{equation*}
and so by the argument appearing in the proof of Lemma~\ref{lem:projection_consistency}, if each ball $B_k$ contains at least one synthetic data point $z_m$, then $\| \hat{\mu}_X - \bar{\mu}_X \|_{\calH} < \varepsilon$. Therefore it suffices to show that if $M \geq C \varepsilon^{-2D(1+a)}$, then the probability of some of the balls not containing any synthetic data point is at most $\gamma$.

To this end, let us look at the number of balls $K$, and the probability that a synthetic data point lands in a particular ball, as functions of $\varepsilon$ (via the ball radius $\frac{\delta}{2}$). First, since $\calX$ is a bounded subset of $\IR^D$, there exists $C_1 \in \IR$ such that for all $\delta > 0$, the space $\calX$ can be covered with $\lfloor C_1 \delta^{-D} \rfloor$ open balls of radii $\delta / 2$. Second, since the density of $q$ is assumed to be bounded away from $0$ on any bounded subset of $\IR^D$, there exists $C_2 \in \IR$ such that $q(B_k) \geq C_2 \delta^D$ for all $k$.

Let $A_k^M$ be the event that the ball $B_k$ remains without a synthetic data point after $M$ of them have been sampled. Then the probability of the event $E_M$ that \emph{any} of the $K$ balls remains empty can be upper bounded by a union bound as
\begin{equation*}
\IP[ E_M ]
\leq \sum_{k = 1}^K \IP[ A_k^M ]
= \sum_{k = 1}^K (1 - q(B_k))^M
\leq \sum_{k = 1}^K (1 - C_2 \delta^D)^M
\leq K \exp\left( - M C_2 \delta^D \right)
\leq C_1 \delta^{-D} \exp\left( - M C_2 \delta^D \right)
.
\end{equation*}
Solving for $M$, we can easily verify that $\IP[E_M] \leq \gamma$ is ensured whenever
\begin{equation*}
M
\geq
\frac{1}{C_2 \delta^D} \left( D \ln \frac{1}{\delta} + \ln \frac{C_1}{\gamma} \right)
=
\frac{(2L)^D}{C_2} \frac{1}{\varepsilon^{2D}} \left( 2 \ln \frac{1}{\varepsilon} + \ln \frac{C_1 (2L)^D}{\gamma} \right)
\end{equation*}
Since $\ln \frac{1}{\varepsilon} < \frac{1}{\varepsilon^a}$ for all sufficiently small $\varepsilon$, we see that we could have chosen $\varepsilon_0 > 0$ and $C \in \IR$ such that the right-hand side is at most $C \varepsilon^{-2D-a}$ for all $\varepsilon \in (0, \varepsilon_0)$. But the condition $M \geq C \varepsilon^{-2D-a}$ is satisfied by supposition, and so we conclude that $\IP\left[ \| \hat{\mu}_X - \bar{\mu}_X \|_{\calH} \right] \leq \IP[E_M] \leq \gamma$.
\end{proof}
\end{lemma}

\begin{hprop}[\ref{prop:alg1_rate_no_publishable_subset}]
Suppose that $\calX$ is a bounded subset of $\IR^D$, the kernel $k$ is Lipschitz, and the synthetic data points $z_1, z_2, \ldots$ are sampled i.i.d.~from a distribution $q$ whose density is bounded away from $0$ on any bounded subset of $\IR^D$. Then $M(N)$ can be chosen so that Algorithm~\ref{alg:synthetic_subspace} outputs an estimator that converges to the true kernel mean embedding $\mu_X$ in RKHS norm at a rate $\calO_p(N^{-1/(D+1+c)})$, where $c$ is any fixed positive number $c > 0$.
\begin{proof}
As in the proof of Theorem~\ref{thm:synthetic_subspace_consistency}, we can decompose the error between the released element $\tilde{\mu}_X$ and the true $\mu_X$ as
\begin{equation}
	\left\| \tilde{\mu}_X - \mu_X \right\|_{\calH}
	\leq
	\underbrace{
		\left\| \hat{\mu}_X - \mu_X \right\|_{\calH}
	}_{\text{finite sample error}}
	+
	\underbrace{
		\left\| \overline{\mu}_X - \hat{\mu}_X \right\|_{\calH}
	}_{\text{projection error}}
	+
	\underbrace{
		\left\| \tilde{\mu}_X - \overline{\mu}_X \right\|_{\calH}
	}_{\text{privacy error}}
	.
\label{app:eq:error_decomposition}
\end{equation}
Using the standard empirical kernel mean embedding estimator, the finite sample error vanishes as $\calO_p(N^{-1/2})$~\citep{muandet_kernel_2016}.
From the proof of Theorem~\ref{thm:synthetic_subspace_consistency} we can see that the privacy error vanishes as $\calO_p(\sqrt{F} / N) \subseteq \calO_p(\sqrt{M} / N)$.
Solving for $\varepsilon$ in the statement of the preceding Lemma~\ref{lem:projection_convergence_rate} we have that for all $\gamma \in (0, 1)$, $a > 0$ and all sufficiently large $M$,
\begin{equation*}
\IP\left[ \| \hat{\mu}_X - \bar{\mu}_X \|_{\calH}
\geq
\frac{1}{C} M^{- \frac{1}{2D+a}} \right]
\leq
\gamma.
\end{equation*}
The projection error thus vanishes at a rate $\calO_p(M^{-1/(2D+a)})$, for any arbitrarily small $a > 0$. To achieve the claimed total rate $\calO_p(N^{-1/(D+1+c)})$ we choose $M(N) = N^k$ with $k = 1 - 4 / (2D + a + 2)$, and verify that
\begin{equation*}
\calO_p\left(
\frac{1}{\sqrt{N}}
+
M^{\frac{-1}{2D+a}}
+
\frac{\sqrt{M}}{N}
\right)
=
\calO_p\left(
\frac{1}{\sqrt{N}}
+
N^{\frac{-k}{2D+a}}
+
\frac{\sqrt{N^k}}{N}
\right)
=
\calO_p\left(
\frac{1}{\sqrt{N}}
+
N^{-\frac{1}{D+1+a/2}}
\right)
=
\calO_p\left(
N^{-\frac{1}{D+1+a/2}}
\right)
\end{equation*}
and the claimed result follows by taking $a = 2c > 0$.
\end{proof}
\end{hprop}

\begin{hprop}[\ref{prop:alg1_rate_publishable_subset}]
Suppose that a fixed proportion $\eta$ of the private database can be published without modification. Using this part of the database as the synthetic data points, Algorithm~\ref{alg:synthetic_subspace} outputs a consistent estimator of $\mu_X$ that converges in RKHS norm at a rate $\calO_p(N^{-1/2})$.
\begin{proof}
Let $\hat{\mu}^{\text{baseline}} := \frac{1}{M} \sum_{m = 1}^M k(z_m, \cdot)$ be the baseline estimator that weights the $M$ public points uniformly. Noting that $\hat{\mu}^{\text{baseline}} \in \calH_M$ lies in the span of feature maps of synthetic data points, for the projection error as defined in equation (\ref{app:eq:error_decomposition}) we have:
\begin{align*}
\left\| \overline{\mu}_X - \hat{\mu}_X \right\|_{\calH}
&=
\left\| \hat{\mu}^{\text{baseline}} - \hat{\mu}_X \right\|_{\calH}
&\text{[ property of projection ]}
\\&=
\left\| \hat{\mu}^{\text{baseline}} - \mu_X \right\|_{\calH}
+
\left\| \hat{\mu}_X - \mu_X \right\|_{\calH}
&\text{[ Triangle inequality ]}
\\&\in
\calO_p\left( M^{-1/2} \right) + \calO_p\left( N^{-1/2} \right)
\end{align*}
Using the error decomposition of equation (\ref{app:eq:error_decomposition}) we thus have
\begin{equation*}
\left\| \tilde{\mu}_X - \mu_X \right\|_{\calH}
\in
\calO_p\left(
N^{-1/2} + (M^{-1/2} + N^{-1/2}) + \sqrt{M} / N
\right)
\end{equation*}
and this is in $\calO_p(N^{-1/2})$ when $M = \eta N$ is proportional to $N$.
\end{proof}
\end{hprop}

\subsection{Algorithm 1 (Synthetic Data Subspace): Differential Privacy}
\label{app:sec:synthetic_subspace_privacy}

The proof of Proposition~\ref{prop:synthetic_subspace_privacy} rests on the following simple calculation:

\begin{lemma}
	\label{lem:KME_RKHS_sensitivity}
	If $k(x, x) \leq 1$ for all $x \in \calX$, then the RKHS norm sensitivity of the empirical kernel mean embedding $\hat{\mu}_X$ with respect to changing one data point is at most $\frac{2}{N}$.
	\begin{proof}
		Let $D = \{ x_1, \ldots ,x_N \}$ and $D' = \{ x'_1, \ldots, x'_N \}$ be two databases of the same cardinality $N$, differing in a single row. Without loss of generality $x_n = x'_n$ for $1 \leq n \leq N - 1$. Let $\hat{\mu}_X$ and $\hat{\mu}_X'$ be the empirical kernel mean embeddings computed using $D$ and $D'$, respectively. Then
		\begin{align}
		\left\| \hat{\mu}_X - \hat{\mu}_X' \right\|_{\calH}
		&=
		\left\| \frac{1}{N} \sum_{n = 1}^N k(x_n, \cdot) - \frac{1}{N} \sum_{n = 1}^N k(x'_n, \cdot) \right\|_{\calH}
		=
		\frac{1}{N} \left\| k(x_N, \cdot) - k(x'_N, \cdot) \right\|_{\calH}
		\\&\leq
		\frac{1}{N} \left(
		\left\| k(x_N, \cdot) \right\|_{\calH} + \left\| k(x_N, \cdot) \right\|_{\calH}
		\right)
		=
		\frac{1}{N} \left(
		k(x_N, x_N)^{1/2} + k(x'_N, x'_N)^{1/2}
		\right)
		\leq
		\frac{2}{N}
		.
		\end{align}
		As $D$ and $D'$ were arbitrary neighbouring databases, the claimed result follows.
	\end{proof}
\end{lemma}

\begin{hprop}[\ref{prop:synthetic_subspace_privacy}.]
	If $k(x, x) \leq 1$ for all $x \in \calX$, then Algorithm~\ref{alg:synthetic_subspace} is $(\varepsilon, \delta)$-differentially private.
	\begin{proof}
		As the synthetic data points $z_1, \ldots, z_M$ do not depend on the private data, it suffices to show that the weights $w_1, \ldots, w_M$ are $(\varepsilon, \delta)$-differentially private. However, these weights result from data-independent post-processing of the coefficients $\boldsymbol{\beta}$, which are a perturbed version of the coefficients $\boldsymbol{\alpha}$, with the perturbation provided by the privacy-protecting \emph{Gaussian mechanism}~\cite{dwork_algorithmic_2014}. It remains to verify that the Gaussian mechanism employs sufficiently scaled noise; in particular we need to verify that $2/N \geq \Delta_2 := \sup_{D, D': D \sim D'} \| \boldsymbol{\alpha} - \boldsymbol{\alpha}' \|_2$.
		
		But indeed, since $b_1, \ldots, b_F$ are orthonormal, for any $\boldsymbol{\alpha}$ and $\boldsymbol{\alpha}'$ computed using neighbouring databases,
		\begin{equation}
		\left\| \boldsymbol{\alpha} - \boldsymbol{\alpha}' \right\|_2
		= \left( \sum_{f = 1}^F (\alpha_f - \alpha'_f)^2 \right)^{1/2}
		= \left\| \overline{\hat{\mu}_N} - \overline{\hat{\mu}'_N} \right\|_{\calH}
		\leq \left\| \hat{\mu}_N - \hat{\mu}'_N \right\|_{\calH}
		\leq \frac{2}{N}
		,
		\end{equation}
		(last inequality is Lemma~\ref{lem:KME_RKHS_sensitivity}) as required to verify the Gaussian mechanism. Then $(\varepsilon, \delta)$-differential privacy for the entire algorithm follows.
	\end{proof}
\end{hprop}

\subsection{Algorithm 2 (Random Features RKHS Algorithm): Consistency}
\label{app:sec:random_features_consistency}

As a preliminary lemma, we first show that a uniform convergence result for the random features $\phi$ translates into a bound on the error incurred by Algorithm~\ref{alg:random_features} due to using random features instead of the original kernel $k$.

\begin{lemma}
Let $\hat{\mu}^{\text{out}}_X := \sum_{m = 1}^M w_m k(z_m, \cdot) \in \calH$ be the element in $\calH$ represented by the output of Algorithm~\ref{alg:random_features}. Let $\hat{\mu}^{\phi, \text{out}}_X := \sum_{m = 1}^M w_m \phi(z_m)$ be the corresponding element in the random features RKHS $\calH_{\phi}$. If the random feature scheme $\phi$ is such that $\sup_{x, x' \in \calX} | \phi(x)^T \phi(x') - k(x, x') | < \delta$, then the following bound on the ``random features error" holds:
\begin{equation*}
\left|
\left\| \hat{\mu}^{\phi, \text{out}}_X - \hat{\mu}^{\phi}_X \right\|_{\calH_{\phi}}
-
\left\| \hat{\mu}^{\text{out}}_X - \hat{\mu}_X \right\|_{\calH}
\right|
\leq
2 \sqrt{\delta}
.
\end{equation*}
\label{app:lem:random_features_uniform_convergence}
\begin{proof}
Expanding the RKHS norms using bilinearity of inner products, we have
\begin{align*}
&\left|
\left\| \hat{\mu}^{\phi, \text{out}}_X - \hat{\mu}^{\phi}_X \right\|_{\calH_{\phi}}
-
\left\| \hat{\mu}^{\text{out}}_X - \hat{\mu}_X \right\|_{\calH}
\right|
\\&=
\Bigg| \left(
\sum_{m_1 = 1}^M \sum_{m_2 = 1}^M w_{m_1} w_{m_2} \phi(z_{m_1})^T \phi(z_{m_2})
+
\sum_{n_1 = 1}^N \sum_{n_2 = 1}^N \frac{1}{N} \frac{1}{N} \phi(x_{n_1})^T \phi(x_{n_2})
-
2 \sum_{m = 1}^M \sum_{n = 1}^N w_m \frac{1}{N} \phi(z_m)^T \phi(x_n)
\right)^{1/2}
\\&-
\left(
\sum_{m_1 = 1}^M \sum_{m_2 = 1}^M w_{m_1} w_{m_2} k(z_{m_1}, z_{m_2})
+
\sum_{n_1 = 1}^N \sum_{n_2 = 1}^N \frac{1}{N} \frac{1}{N} k(x_{n_1}, x_{n_2})
-
2 \sum_{m = 1}^M \sum_{n = 1}^N w_m \frac{1}{N} k(z_m, x_n)
\right)^{1/2} \Bigg|
\end{align*}
Since $\sum_{m = 1}^M |w_m| \leq 1$ by construction and $\sum_{n = 1}^N \frac{1}{N} = 1$, thanks to the assumption on $\phi$ this expression is of the form
\begin{equation*}
\Big| (a + b + 2c)^{1/2} - (A + B + 2C)^{1/2} \Big|
\end{equation*}
for suitable $a, A, b, B, c, C \in \IR$ with $|a - A|, |b - B|, |c - C| < \delta$. By monotonicity of the square root function, this expression is maximised when $A = a + \delta$, $B = b + \delta$, $C = c + \delta$. Writing $s := a + b + 2C$, we have
\begin{equation*}
\Big| (a + b + 2c)^{1/2} - (A + B + 2C)^{1/2} \Big|
\leq | s^{1/2} - (s + 4 \delta)^{1/2} |
= (s + 4 \delta)^{1/2} - s^{1/2}
\leq s^{1/2} + 2 \delta^{1/2} - s^{1/2}
= 2 \delta^{1/2}
.
\qedhere
\end{equation*}
\end{proof}
\end{lemma}

\begin{hthm}[\ref{thm:random_features_consistency}]
	\label{app:thm:random_features}
	Suppose that the random features $\phi$ converge $\phi(\cdot)^T \phi(\cdot) \to k(\cdot, \cdot)$ uniformly in $\calX$ as the number of random features $J \to \infty$. Assume also availability of an approximate Reduced set construction method that solves the minimisation (\ref{eq:random_features_preimage_minimisation}) either up to a constant multiplicative error, or with an absolute error that can be made arbitrarily small. Then Algorithm~\ref{alg:random_features} outputs a consistent estimator of the kernel mean embedding $\mu_X$ in the sense that
	\begin{equation}
	\sum_{m = 1}^M w_m k(z_m, \cdot)
	\stackrel{\IP}{\to}
	\mu_X
	\hspace{2em} \text{as } N \to \infty
	.
	\end{equation}
	\begin{proof}
		The output of Algorithm~\ref{alg:random_features} specifies an element $\hat{\mu}^{\text{out}}_X := \sum_{m = 1}^M w_m k(z_m, \cdot) \in \calH$ in the RKHS $\calH$ of $k$. Its RKHS distance to the true kernel mean embedding $\mu_X$ of $X$ can be upper bounded by a decomposition using the Triangle inequality, where we write $\hat{\mu}^{\phi, \text{out}}_X := \sum_{m = 1}^M w_m \phi(z_m)$ for the element of $\calH_{\phi}$ that the Reduced set method constructs to approximate the privacy-protected $\tilde{\mu}^{\phi}_X$:
		\begin{align}
		\left\| \mu_X - \hat{\mu}^{\text{out}}_X \right\|_{\calH}
		&\leq
		\underbrace{
			\left\| \mu_X - \hat{\mu}_X \right\|_{\calH}
		}_{\text{finite sample error}}
		+
		\underbrace{
			\left\| \hat{\mu}_X - \hat{\mu}^{\text{out}}_X \right\|_{\calH}
		}_{\text{other errors}}
		\nonumber\\&\leq
		\underbrace{
			\left\| \mu_X - \hat{\mu}_X \right\|_{\calH}
		}_{\text{finite sample error}}
		+
		\underbrace{
			\left| \left\| \hat{\mu}^{\phi, \text{out}}_X - \hat{\mu}^{\phi}_X \right\|_{\calH_{\phi}} - \left\| \hat{\mu}_N - \hat{\mu}^{\text{out}}_X \right\|_{\calH} \right|
		}_{\text{random features error}}
		+
		\underbrace{
			\left\| \hat{\mu}^{\phi, \text{out}}_X - \hat{\mu}^{\phi}_X \right\|_{\calH_{\phi}}
		}_{\text{other errors}}
		\nonumber\\&\leq
		\underbrace{
			\left\| \mu_X - \hat{\mu}_X \right\|_{\calH}
		}_{\text{finite sample error}}
		+
		\underbrace{
			\left| \left\| \hat{\mu}^{\phi, \text{out}}_X - \hat{\mu}^{\phi}_X \right\|_{\calH_{\phi}} - \left\| \hat{\mu}_N - \hat{\mu}^{\text{out}}_X \right\|_{\calH} \right|
		}_{\text{random features error}}
		\nonumber\\&+
		\underbrace{
			\left\| \hat{\mu}^{\phi, \text{out}}_X - \tilde{\mu}^{\phi}_X \right\|_{\calH_{\phi}}
		}_{\text{reduced set error}}
		+
		\underbrace{
			\left\| \tilde{\mu}^{\phi}_X - \hat{\mu}^{\phi}_X \right\|_{\calH_{\phi}}
		}_{\text{privacy error}}
		.
		\label{app:eq:alg2_error_decomposition}
		\end{align}
		The finite sample error tends to $0$ as $N \to \infty$ in probability by consistency of the empirical kernel mean estimate. The random features error goes to $0$ as $N \to \infty$ by Lemma~\ref{app:lem:random_features_uniform_convergence}, since $J \to \infty$ as $N \to \infty$ and $\phi(\cdot)^T \phi(\cdot) \to k(\cdot, \cdot)$ uniformly in $\calX$ as $J \to \infty$. The privacy error goes to $0$ as $N \to \infty$ by the same argument as in the proof of Theorem~\ref{thm:synthetic_subspace_consistency}, with $F$ replaced by $J$. So it remains to show that the reduced set error also goes to $0$ as $N \to \infty$, in probability.
		
		First, note that the private empirical kernel mean embedding $\hat{\mu}_X^{\phi} = \frac{1}{N} \sum_{n = 1}^N \phi(x_n)$ is in the feasible set of the constrained minimisation problem solved by the reduced set method, as the sum of absolute values of weights in this expansion is $N | \frac{1}{N} | = 1 \leq 1$. The RKHS $\calH_{\phi}$ distance of $\hat{\mu}_X^{\phi}$ to the optimisation target $\tilde{\mu}^{\phi}_X$ equals the privacy error, so it follows that the reduced set error is upper bounded by the privacy error, and hence also goes to $0$ as $N \to \infty$:
		\begin{equation}
		\underbrace{
			\left\| \hat{\mu}^{\phi, \text{out}}_X - \tilde{\mu}^{\phi}_X \right\|_{\calH_{\phi}}
		}_{\text{reduced set error}}
		\leq
		\underbrace{
			\left\| \tilde{\mu}^{\phi}_X - \hat{\mu}^{\phi}_X \right\|_{\calH_{\phi}}
		}_{\text{privacy error}}
		\stackrel{\IP}{\to} 0 \text{ as } N \to \infty
		,
		\end{equation}
		as required to complete the proof.
	\end{proof}
\end{hthm}

\begin{corollary}
	Let $f$ be any continuous function. Then whenever $k$ is a $c_0$-universal kernel, applying $f$ to the points output by Algorithm~\ref{alg:random_features} yields a consistent estimator of the kernel mean embedding $\mu_{f(X)}$ of $f(X)$.
	\begin{proof}
		Noting that the sum of absolute values of weights $w_m$ output by Algorithm~\ref{alg:random_features} is at most $C$ by construction, in light of Theorem~\ref{thm:random_features_consistency} we see that Theorem 1 of~\cite{scibior_consistent_2016} applies and gives the desired conclusion.
	\end{proof}
\end{corollary}

\subsection{Algorithm 2 (Random Features RKHS Algorithm): Convergence Rate}
\label{app:sec:random_features_convergence_rate}

\begin{hprop}[\ref{thm:random_features_consistency}]
Suppose that $\phi$ is a random feature scheme for the kernel $k$ that converges uniformly on any compact set at a rate $\calO_p(J^{-1/2})$ with the number $J$ of random features. Then $J(N)$ can be chosen such that if the employed Reduced set method finds a global optimum of (\ref{eq:random_features_preimage_minimisation}), Algorithm~\ref{alg:random_features} outputs an element that converges to the true kernel mean embedding $\mu_X$ at a rate $\calO_p(N^{-1/3})$.
\begin{proof}
Equation (\ref{app:eq:alg2_error_decomposition}) shows that the error $\left\| \mu_X - \hat{\mu}^{\text{out}}_X \right\|_{\calH}$ between the released element $\hat{\mu}^{\text{out}}_X$ and the true kernel mean embedding $\mu_X$ can be upper bounded by the sum of four terms: the finite sample error, the random features error, the reduced set error, and the privacy error.
Arguing as in the proof of Proposition~\ref{prop:alg1_rate_no_publishable_subset}, the finite sample error vanishes at a rate $\calO_p(N^{-1/2})$.
The proof of Theorem~\ref{thm:random_features_consistency} shows that the reduced set error is upper bounded by the privacy error, which itself vanishes at a rate of $\calO_p(\sqrt{J}/N)$ by the argument given in the proof of Theorem~\ref{thm:synthetic_subspace_consistency}, with $F$ replaced by $J$.
Lemma~\ref{app:lem:random_features_uniform_convergence} implies that if the random features converge uniformly at a rate $\calO_p(J^{-1/2})$, then the random features error vanishes at a rate $\calO_p(J^{-1/4})$. The total convergence rate is thus
\begin{equation*}
\calO_p\left( N^{-1/2} + \frac{\sqrt{J}}{N} + J^{-1/4} \right)
\end{equation*}
and we can check that this becomes $\calO_p(N^{-1/3})$ by setting $J = \lfloor N^{4/3} \rfloor$.
\end{proof}
\end{hprop}

\subsection{Algorithm 2 (Random Features RKHS Algorithm): Differential Privacy}
\label{app:sec:random_features_privacy}

\begin{hprop}[\ref{prop:random_features_privacy}]
	Assume that the random feature vectors produced by $\phi$ are bounded by $1$ in $L_2$ norm ($\| \phi(x) \|_2 \leq 1$ for all $x \in \calX$). Then Algorithm~\ref{alg:random_features} is $(\varepsilon, \delta)$-differentially private.
	\begin{proof}
		The output of the algorithm is produced by a Reduced set method that is initialised blindly to the database and optimises RKHS distance to the element $\tilde{\mu}^{\phi}_X \in \calH_{\phi}$, while only having access to the distance to it, rather than any representation of $\tilde{\mu}^{\phi}_X$. As $\tilde{\mu}^{\phi}_X$ can be seen as a vector in $\IR^J$ obtained by perturbing $\hat{\mu}^{\phi}_X$ using the Gaussian mechanism with $\Delta_2 = \frac{2}{N}$, it suffices to show that the $L_2$-sensitivity of $\hat{\mu}^{\phi}_X$ is upper bounded by $\frac{2}{N}$. To this end, assume $D = \{ x_1, \ldots, x_N \}$ and $D' = \{ x'_1, \ldots, x'_N \}$ are two neighbouring databases of cardinality $N$, differing w.l.o.g.~in their last element only. Then
		\begin{align}
		\| \hat{\mu}_{D}^{\phi} - \hat{\mu}_{D'}^{\phi} \|_2
		&=
		\Bigg\| \frac{1}{N} \sum_{n = 1}^N \phi(x_n) - \frac{1}{N} \sum_{n = 1}^N \phi(x_n') \Bigg\|_2
		\\&=
		\frac{1}{N} \| \phi(x_N) - \phi(x_N') \|_2
		\\&\leq
		\frac{1}{N} \| \phi(x_N) \|_2 + \frac{1}{N} \| \phi(x_N') \|_2
		\leq
		\frac{2}{N}
		,
		\end{align}
		as required to complete the proof.
	\end{proof}
\end{hprop}

\section{Setup of Empirical Illustrations}
\label{app:sec:experiments}

We considered two scenarios in our basic empirical evaluations shown in Sections~\ref{sec:perturb_in_synthetic_subspace} and~\ref{sec:perturb_in_random_features_RKHS}:
\begin{enumerate}
	\item \emph{No publishable subset}: No rows of the private database are, or can be made public without some privacy-ensuring modification.
	\item \emph{Publishable subset}: A small part of the private database is already public, or can be made public, perhaps for one of the several possible reasons outlined in Section~\ref{sec:introduction}.
\end{enumerate}
To illustrate the impact of data dimensionality on the performance of the proposed algorithms, we provide results on datasets with data dimension $D = 2$ and $D = 5$. In both cases we constructed a synthetic private dataset by sampling $N = 100,000$ data points from a multivariate Gaussian mixture distribution. The mixture had $10$ components, with mixing weights proportional to $1, \frac{1}{2}, \ldots, \frac{1}{10}$, and the means of the components were chosen randomly themselves from a spherical Gaussian distribution with mean $[100, \ldots, 100]$ and covariance $200 I_D$. Each of the $N$ private data points was simulated by first sampling its mixture component using the mixing weights as probabilities, and then the point itself was sampled from a spherical Gaussian centered at the mean of the chosen mixture component and with covariance $30 I_D$.

We chose to work with the widely popular exponentiated quadratic kernel $k(x_1, x_2) = e^{-\gamma \| x_1 - x_2 \|_2^2}$ for $\IR^D$-valued data (also known as a Gaussian kernel, or a squared exponential kernel), with the parameter setting $\gamma = 10^{-4} / D$. This kernel is known to be \emph{characteristic}~\cite{FukGreSunSch08}, and so as discussed in Section~\ref{sec:background:kernels}, no information about the data generating distribution $p_X$ is lost by working with its kernel mean embedding $\mu_X$.

We used our proposed algorithms to release an approximate version of the empirical KME of the private database, in such a way that the output satisfies the definition of $(\varepsilon, \delta)$-differential privacy. We investigated the common privacy levels given by $\varepsilon \in \{ 0.01, 0.1, 1.0 \}$, and used the fixed value of $\delta = 10^{-6}$, which satisfies the usual requirement that $\delta \ll \frac{1}{N}$.

\subsection{Evaluation Metric}

The geometry of the RKHS $\calH$ allows comparing the performance of different algorithms by computing the RKHS distance $\Delta$ between the empirical KME $\hat{\mu}_X$ computed using all $N$ private data points (and which could not have been released without violating differential privacy) and the element of the RKHS represented by the actually released weighted set of synthetic data points $(z_1, w_1), \ldots, (z_M, w_M)$:
\begin{equation*}
\Delta := \left\| \hat{\mu}_X - \sum_{m = 1}^M w_m k(z_m, \cdot) \right\|_{\calH}
.
\end{equation*}
Moreover, as the empirical KME $\hat{\mu}_X$ is based on a large sample size of $N = 100,000$ i.i.d.~data points, it can be expected to be a good proxy for the true KME $\mu_X$ of the data-generating random variable $X$. In that case $\Delta$ is also a good proxy for the RKHS distance between the true KME $\mu_X$ and the RKHS element represented by the released dataset.

\subsection{Scenario 1: No Publishable Subset}
\label{sec:experiments:nodata}

Algorithm~\ref{alg:synthetic_subspace} requires specifying the synthetic data points $z_1, \ldots, z_M$ in advance, before seeing the private data. If no part of the private data has already been published (which could then be used for the synthetic data points), one can construct the synthetic data points by sampling them randomly from a suitable probability distribution $q$. For the consistency result of Theorem~\ref{thm:synthetic_subspace_consistency} to apply, the support of $q$ must include all possible private data points. In our case the private data takes values in $\IR^D$, and so this requirement is satisfied by any distribution on $\IR^D$ with full support. We used a spherical Gaussian distribution $q = \mathcal{N}(0, \sigma_q I_D)$ with $\sigma_q = 500$ for sampling the synthetic data points.

The implementation of Algorithm~\ref{alg:random_features} used $J = 10,000$ random features for accurate approximation of the kernel, and an iterative gradient-based optimisation procedure to solve the reduced set problem (Equation (\ref{eq:random_features_preimage_minimisation}) in Algorithm~\ref{alg:random_features}).

Figure~\ref{fig:nodata} shows how the RKHS distance $\Delta$ changes with the number of synthetic data points $M$, for different requested privacy level $\varepsilon$ for Algorithm~\ref{alg:synthetic_subspace} (solid lines) and Algorithm~\ref{alg:random_features} (dashed lines), on datasets with dimensionality $D = 2$ (left subfigure) and $D = 5$ (right subfigure). We observe that the additional ability of Algorithm~\ref{alg:random_features} to optimise the \emph{locations} of the synthetic data points (rather than just the weights, as is the case for Algorithm~\ref{alg:synthetic_subspace}) is more helpful in the higher-dimensional case $D = 5$, where the randomly sampled synthetic data points are less likely to land close to private data points.

\subsection{Scenario 2: Publishable Subset}
\label{sec:experiments:leaks}

Here we explored the interesting scenario where one can exploit the fact that a small part of the private database is actually public, and use the public rows as the fixed synthetic data points in Algorithm~\ref{alg:synthetic_subspace}. Specifically, we assume (without loss of generality) that the first $M$ rows of the private database (where $M \ll N$) are public, and we take the synthetic data points to be $z_1 \gets x_1, \ldots, z_M \gets x_M$.

Observe that in this case $\hat{\mu}^{\text{baseline}} := \frac{1}{M} \sum_{m = 1}^M k(z_m, \cdot)$, i.e., uniform weighting of the synthetic data points, is already expected to be a strong baseline since $\hat{\mu}^{\text{baseline}}$ is itself a consistent estimator of $\mu_X$, (although based on a much smaller sample size $M \ll N$). The purpose of Algorithm~\ref{alg:synthetic_subspace} is to find (generally non-uniform) $w_1, \ldots, w_M$ that reweight the public data points using the information in the large private dataset, but respecting differential privacy. Figure~\ref{fig:leaks} shows how the RKHS distance $\Delta$ changes with the number of public data points $M$, for different privacy levels $\varepsilon$.

For comparison, the figures also show the RKHS distances $\Delta$ achieved by the baseline that simply weights the public points uniformly. We can see that in both cases $D = 2$ and $D = 5$, if the ratio of public to private points is low enough, Algorithm~\ref{alg:synthetic_subspace} provides a substantial benefit over this baseline (note the logarithmic scaling). Since usually obtaining permission to publish a larger subset of the private data unchanged will come at an increased cost, the ability to instead reweight a smaller public dataset using Algorithm~\ref{alg:synthetic_subspace} in a differentially private manner is useful.

\end{document}